%% file: root.tex
\documentclass[letterpaper, 10 pt, journal, twoside]{IEEEtran}
\ifCLASSINFOpdf
\else
\fi
\usepackage{amsmath}
\usepackage{amssymb}
\usepackage{bm}
\usepackage{subcaption}
\usepackage{hyperref}
\usepackage{graphicx}
\usepackage{xcolor}
\usepackage{caption}
\usepackage[font=footnotesize]{caption}
\usepackage{tikz}
\definecolor{mycolor}{rgb}{0,0.5,0.3}
\usepackage{pdfpages}
\usepackage{makecell,booktabs,multirow} 
\hypersetup{
    colorlinks=true,
    linkcolor=blue,
    filecolor=magenta,      
    urlcolor=cyan,
}
\newcommand{\pseudoparagraph}[1]{\paragraph{#1} }

\makeatletter
\newcommand\notsotiny{\@setfontsize\notsotiny\@vipt\@viipt}
\makeatother

\newsavebox\CBox
\def\textBF#1{\sbox\CBox{#1}\resizebox{\wd\CBox}{\ht\CBox}{\textbf{#1}}}

\newcolumntype{x}[1]{>{\centering\arraybackslash\hspace{0pt}}p{#1}}

\newcommand\blfootnote[1]{%
   \begingroup
   \renewcommand\thefootnote{}\footnote{#1}%
   \addtocounter{footnote}{-1}%
   \endgroup
}

\begin{document}
%
\title{E-NeRF: Neural Radiance Fields from a Moving Event Camera}
%
%
%

\author{Simon Klenk$^{1}$, Lukas Koestler$^{1}$, Davide Scaramuzza$^{2}$, Daniel Cremers$^{1}$
\thanks{Manuscript received August 19, 2022; revised December 6, 2022; Accepted: January 13, 2023.} 
\thanks{This paper was recommended for publication by Editor Javier Civera upon evaluation of the Associate Editor and Reviewers' comments. 
This work was supported by the ERC Advanced Grant SIMULACRON, the Munich Center for Machine Learning, the BMBF Project BBKI-Chips, the Swiss National Science Foundation (SNSF) through the National Centre of Competence in Research (NCCR) Robotics, and the European Research Council (ERC) under Grant Agreement 864042 (AGILEFLIGHT).}
\thanks{$^{1}$First, Second and Fourth Author are with Computer Vision Group, Technical University of Munich, Germany.\ $^{2}$Third author is with Robotics and Perception Group, University of Zurich, Switzerland}
\thanks{Digital Object Identifier (DOI): see top of this page.}
}

%
%

\markboth{IEEE Robotics and Automation Letters. Preprint Version. Accepted January, 2023}
{Klenk \MakeLowercase{\textit{et al.}}: E-NeRF: Neural Radiance Fields from a Moving Event Camera} 

%



\twocolumn[{
\renewcommand\twocolumn[1][]{#1}%
\maketitle
\centering
\vspace{0.5cm}
\includegraphics[clip, trim=0cm 17.7cm 0cm 0.8cm, width=\textwidth]{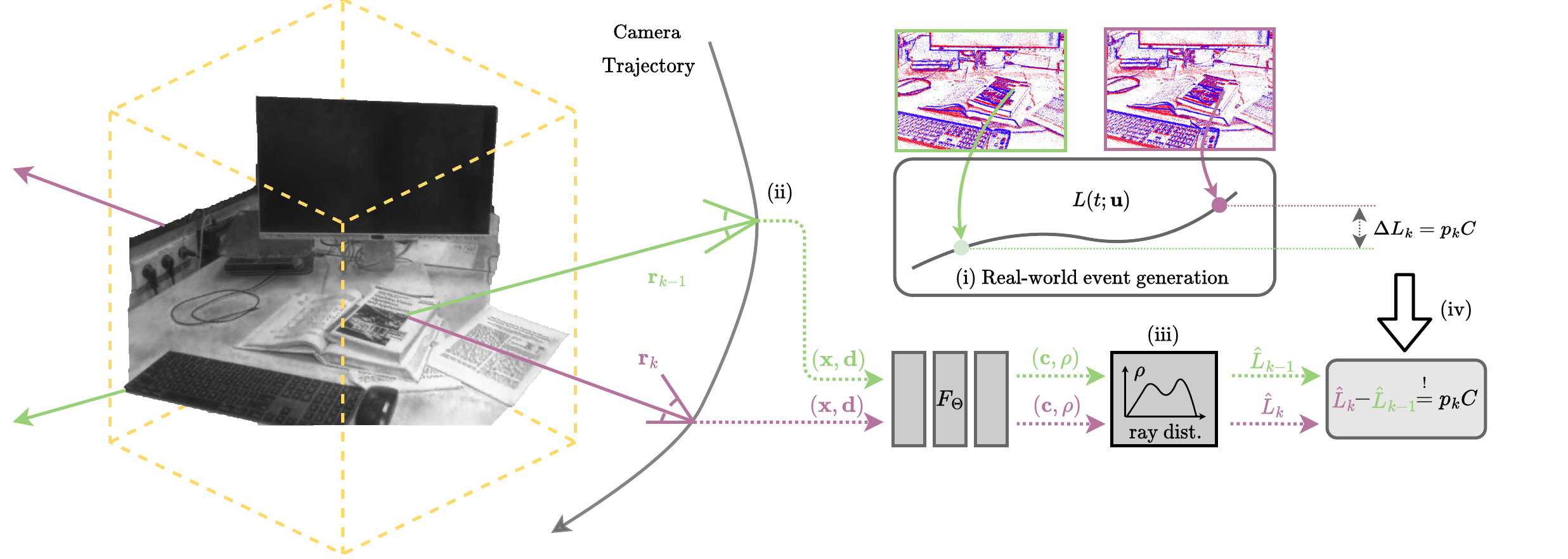}
\vspace{2ex}
\captionof{figure}{\label{fig::overview}Overview of the proposed E-NeRF. (i) For a fixed pixel $\mathbf{u} \in \mathbb{R}^2$ the real-world logarithmic brightness $L(t; \mathbf{u})$ varies over time. If $L(t; \mathbf{u})$ crosses a threshold $C$ wrt.\ to the memory value a new event is triggered (cf.~\autoref{eq:evModel}). (ii) The proposed E-NeRF reconstructs a radiance field from a moving event camera. We evaluate the NeRF-MLP $F_{\Theta}$ at the two rays $\mathbf{r}_{k-1}$ and $\mathbf{r}_{k}$ corresponding to the two last events $e_{k-1}$ and $e_k$ for pixel $\mathbf{u}$ to obtain color and density $(\mathbf{c}, \rho)$. (iii) The color and density for many points along each ray are accumulated into the final estimated pixel brightness $\hat{I}_{k-1}$ and $\hat{I}_{k}$ using volume rendering. We convert the brightness with the linlog mapping in \autoref{eq:linlog} to $\hat{L}_{k-1}$ and $\hat{L}_k$. (iv) The event constraint induces the loss function on the two rendered logarithmic brightness values, see \autoref{eq:evModel2} (and \autoref{eq:evModel3} for unknown threshold $C$).
\vspace{4ex}
}
}]

\blfootnote{Manuscript received August 19, 2022; revised December 6, 2022; Accepted: January 13, 2023. \newline
\indent This paper was recommended for publication by Editor Javier Civera upon evaluation of the Associate Editor and Reviewers' comments. \newline
\indent This work was supported by the European Research Council (ERC) under Grant Agreements 884679 (SIMULACRON) and 864042 (AGILEFLIGHT), the Munich Center for Machine Learning, the BMBF Project BBKI-Chips, the Swiss National Science Foundation (SNSF) through the National Centre of Competence in Research (NCCR) Robotics.\newline
\indent  $^{1}$The first, second, and fourth authors are with Computer Vision Group, Technical University of Munich, Germany.\ $^{2}$The third author is with the Robotics and Perception Group, University of Zurich, Switzerland.\newline
\indent Digital Object Identifier (DOI): see top of this page.}


\vspace{-0.37cm}
\begin{abstract}
Estimating neural radiance fields (NeRFs) from "ideal" images has been extensively studied in the computer vision community. Most approaches assume optimal illumination and slow camera motion. These assumptions are often violated in robotic applications, where images may contain motion blur, and the scene may not have suitable illumination. This can cause significant problems for downstream tasks such as navigation, inspection, or visualization of the scene. To alleviate these problems, we present E-NeRF, the first method which estimates a volumetric scene representation in the form of a NeRF from a fast-moving event camera. Our method can recover NeRFs during very fast motion and in high-dynamic-range conditions where frame-based approaches fail. We show that rendering high-quality frames is possible by only providing an event stream as input. Furthermore, by combining events and frames, we can estimate NeRFs of higher quality than state-of-the-art approaches under severe motion blur. We also show that combining events and frames can overcome failure cases of NeRF estimation in scenarios where only a few input views are available without requiring additional regularization.
\end{abstract}


%
\IEEEpeerreviewmaketitle

\hfill \break
\noindent
Code and simulated datasets can be found under \url{https://github.com/knelk/enerf}.

\section{Introduction}
%
%
%
%
\IEEEPARstart{C}{ameras} are frequently used in robotic perception for tasks such as localization, path planning, scene understanding, and inspection.
Furthermore, cameras are ubiquitous in virtual and augmented reality systems, where they are used for ego localization, spatial reasoning, and visualizations. 
These systems require compact and high-quality visual representations of the underlying three-dimensional geometry. 

Recently, there has been a trend in the computer vision community to study neural radiance fields (NeRFs) as a solution to scene representation and novel view synthesis. NeRFs represent the scene by a multilayer perceptron (MLP) combined with differentiable rendering \cite{mildenhall2020nerf}. They allow for novel view synthesis at unseen viewpoints. Up to this date, NeRFs have mainly been studied on simulated data as well as on high-quality, real-world images gathered under ideal conditions \cite{nerfSurvey1, nerfSurvey2}, which ensure good surrounding light and minimal noise, apart from a few exceptions such as \cite{ma2021deblur, mildenhall2021rawnerf}. 

One common problem of real-world images is motion blur, which is caused by fast motion and high exposure times (usually required in low-light scenes). During the exposure time interval, each pixel of the moving camera integrates light from different points in the scene, resulting in a mixture of color values. Hence, motion blur is essentially a loss of information that can cause robotic navigation pipelines to fail \cite{robusteventVO22, ulimateSlam18} and severely limit downstream processing, e.g., inspection tasks. 
Simply reducing the exposure time is often not feasible since this also reduces the amount of light received on the sensor and increases the amount of noise. 
Another problem of real-world images from conventional cameras is their limited dynamic range, which can cause bright regions of a scene to appear fully white and dark regions to appear fully black. This is a problem if the affected image areas contain important information, such as text.

The event camera \cite{gallego2020Survey} is a type of camera that addresses both these shortcomings of conventional cameras. Hence, it can replace and complement frame-based cameras in tasks where high dynamic range (HDR) and fast motion are common. Motivated by this fact, we propose to use an event camera for estimating NeRFs in challenging real-world capturing conditions. The main contributions of this paper are: 
\begin{itemize}
    \item 
    This is the first attempt to tackle NeRF estimation from event cameras during very fast motion.
    \item Our method can combine the event stream with (color) images. This allows estimating NeRFs in challenging conditions where frame-based approaches fail or require additional regularization, e.g., under strong motion blur or if only a few frame-based views are available.
    \item We open-source our code and the simulated datasets. 
\end{itemize}


%

\section{Related Work}
This work tackles the problem of reconstructing a 3D neural radiance field (NeRF) from a moving event camera. We thus present related work that reconstructs NeRFs in challenging scenarios and methods that perform 2D or 3D reconstruction from event data. We refer interested readers to the excellent resources on neural rendering \cite{nerfSurvey1,nerfSurvey2}, event-based vision \cite{gallego2020Survey}, and 3D reconstruction \cite[Chap.~13]{DBLP:series/txcs/Szeliski22}.

\pseudoparagraph{3D Scene Reconstruction}
The reconstruction of 3D scenes from a moving RGB camera has traditionally been tackled by methods that produce sparse point clouds \cite{dsoEngelpami2018, schoenberger2016sfm}, depth maps~\cite{Newcombe2011dtam} or voxel grids~\cite{koestler2021tandem}. Recently, NeRFs~\cite{mildenhall2020nerf} have gained popularity due to their ability to synthesize high-quality novel views.
Usually, the input images to NeRF contain little noise. 
One exception to this is Deblur-NeRF~\cite{ma2021deblur} which handles blurry input images by modeling the blur formation. Their method is limited to moderate motion blur, and they only model nonconsistent blur, i.e. they can only recover a NeRF for blur that is caused by irregular (non-straight, non-circular) motion trajectories. RawNeRF \cite{mildenhall2021rawnerf} can deal with images captured under low light conditions. Their method requires raw images and does not handle motion blur, which makes it applicable only to scenes captured with a temporarily static camera. Overall, none of these methods work for scenes captured by a moving camera under low light conditions, however, these conditions are common in robotics applications. 


\pseudoparagraph{Event Cameras}
Event cameras are vision sensors which transmit binary events per pixel, indicating an increase or decrease of the observed brightness. Converting this asynchronous data stream to images is well-studied in 2D \cite{Rebecq19pami, Pan20pami, Scheerlinck18accv, BardwoCVPR16}. Rebeq et al.~\cite{Rebecq19pami} train a recurrent U-Net on synthetic events and reconstruct intensities of high speed phenomna. However, since the event stream does not contain complete photometric information (it only contains derivative data and isomotion-dependent), several approaches combine frames with events. For example, Tulyakov et al. \cite{Tulyakov21cvpr} train a fusion network for color frames and events, showing impressive results on the task of frame rate upsampling.
Without requiring any training data, the approach by Pan et al.~\cite{Pan20pami} combines motion-blurred frames with events using energy-based minimization. They model motion blur as an integral over the exposure time, and optimize for a global event threshold.  Similarly, the work by Scheerlinck et al. \cite{Scheerlinck18accv} fuses frames with events using a classical filtering technique. Zhou et al.~\cite{gallegoSemiDense3d18eccv} compute a semi-dense 3D reconstruction from event data with known poses, as well as by estimating the poses simultaneously  \cite{yiZhou21EVO, hidalgo2022event}. These works produce sparse point clouds for navigation purposes, which in contrast to our work are not well-suited for (novel-view) visualizations.  

\input{tables/spiral}

\pseudoparagraph{Concurrent Work}
Rudnev et al.~\cite{rudnev2022eventnerf} (EventNeRF) and Hwang et al.~\cite{evNerf} (Ev-NeRF) investigate how NeRFs could be reconstructed from event data. These works show the need of the event vision community to generate 3D reconstructions from events. 
Rudnev et al.~\cite{rudnev2022eventnerf} aim to reconstruct visually pleasing NeRFs in color space by using a color event camera. Their setting is constrained to a static event camera with controlled illumination and fully controlled object motion (rotation on a turntable). They assume that events are only triggered by the foreground object, and require the background color to be known.
In contrast, we focus on estimating NeRFs from a moving event camera with many background events and under general illumination. 
Hwang et al.~\cite{evNerf} reconstruct NeRFs from a slowly moving event camera without considering motion blur.
In contrast to both of these works, we focus our evaluation on failure cases of frame-based cameras. Additionally, we are the only work to show NeRF reconstructions from a combination of events and frames leveraging the advantages of both sensor modalities. We introduce a novel no-event loss function and we do not require the contrast threshold to be determined explicitly. Furthermore, E-NeRF does not rely on an integration window but employs directly neighboring events in the loss.

\section{Method}
\subsection{Optimizing a Neural Radiance Field}
Akin to NeRF \cite{mildenhall2020nerf}, E-NeRF optimizes a neural radiance field parameterized by an MLP $F_{\Theta}: (\mathbf{x}, \mathbf{d}) \to (\mathbf{c}, \rho)$, which maps Cartesian input coordinates $\mathbf{x} \in \mathbb{R}^3 $ and viewing direction $\mathbf{d} \in \mathbb{S}^2$ to the predicted color $\mathbf{c} \in \mathbb{R}^3$ and volume density $\rho \in \mathbb{R}$, see \autoref{fig::overview}. The predicted color value $I(\mathbf{u})$ at pixel $\mathbf{u} \in \mathbb{R}^2$ is obtained by volumetric rendering~\cite{DBLP:journals/tvcg/Max95a} of $N$ color predictions $\{\mathbf{c}_{k,j}\}^{j=N}_{j=1}$ along the ray $\mathbf{r}_k$, where $\mathbf{r}_k = \pi^{-1}(\mathbf{T}(t_k), \mathbf{u}_k, \mathbf{K}, \bm{\kappa})$, with interpolated camera poses $\mathbf{T}(t_k) \in SE(3)$, pre-calibrated intrinsics $\mathbf{K}$ and pre-calibrated distortion parameters $\bm{\kappa}$. We perform uniform sampling along each ray. The original NeRF minimizes a least squares error between the rendered predictions $\hat{I}(\mathbf{u})$ and ground truth colors $I(\mathbf{u})$ provided by the images, whereas E-NeRF utilizes the event generation model \autoref{eq:evModel} to optimize $F_{\Theta}$, which is detailed in the next section. 
Because ordinary neural networks do not accurately represent high-frequency details, one key implementation detail is to use a proper encoding \cite{TancikNIPS20TangentKernel} of input coordinates $(\mathbf{x}, \mathbf{d})$. NeRF solves this by applying a Fourier-based feature mapping, whereas E-NeRF use a multi-resolution hash-encoding~\cite{mueller2022instant}.

\subsection{Event Stream Utilization}
Input to our optimization problem is a continuous stream of events $e_k = (\mathbf{u}_k, t_k, p_k)$ which occur asynchronously at pixel $\mathbf{u}_k$ with micro-second timestamp $t_k$. The polarity $p_k \in \{+1, -1\}$ indicates an increase or decrease of the logarithmic brightness $L(\mathbf{u}_k, t_k)$ by the contrast threshold $C$, i.e. an event at time $t_k$ is triggered if the following condition holds:

\begin{equation}\label{eq:evModel}
\Delta L = L(\mathbf{u}_k, t_k) - L(\mathbf{u}_k, t_{k-1}) = p_k \, C
\end{equation} where $t_{k-1}$ is the time of the last event at pixel $\mathbf{u}_k$ and 

\begin{equation}\label{eq:linlog}
    L(\mathbf{u}, t) = \text{linlog}(I(\mathbf{u})) = 
    \begin{cases}
    I(\mathbf{u}) \cdot ln(B) / B, & \text{if}\ I(\mathbf{u}) < B\\
    ln(I(\mathbf{u})), & \text{else}.
  \end{cases}
\end{equation} The threshold $B$ determines the linear region, where no logarithmic mapping is applied. Following v2e \cite{v2e} we set $B = 20$, which models realistic event distributions. For each event $e_k$, we compute two associated rays $\mathbf{r}_k$ and $\mathbf{r}_{k-1}$ at time $t_k$ and $t_{k-1}$, respectively, see \autoref{fig::overview}. We query the MLP for its color predictions $\{\mathbf{c}_{k,j}\}^{j=N}_{j=1}$ and $\{\mathbf{c}_{k-1,j}\}^{j=N}_{j=1}$ along the respective rays and perform volumetric rendering to obtain $\hat{I}(\mathbf{u}_k)$ and $\hat{I}(\mathbf{u}_{k-1})$. We then perform the linlog mapping in equation \ref{eq:linlog} to obtain the predicted logarithmic brightness difference $\Delta \hat{L}_k = \hat{L}(\mathbf{x}_k, t_k) - \hat{L}(\mathbf{x}_k, t_{k-1})$. 
Stacking all $N_{\text{evs}}$ predictions into $\Delta \hat{\mathbf{L}} \in \mathbb{R}^{N_\text{evs}}$ and assuming a Gaussian distribution of residuals $\Delta \hat{L}_k - \Delta L_k$, we perform a least squares minimization of the event loss 
\begin{equation}\label{eq:evModel2}
\mathcal{L}_\text{evs}(\Theta)
 = \|
\Delta \hat{\mathbf{L}}(\Theta) - \Delta \mathbf{L}(\Theta) \|_2^2,
\end{equation}
where $\Delta L_k = p_k \, C$ is measured by the event camera. The contrast threshold $C$ of a real event camera varies over the image plane and over time  \cite{v2e}, which can make \autoref{eq:evModel} impractical to use in a real-world setup. Hence, inspired by~\cite{hidalgo2022event}, for real-world data we propose to use
\begin{equation}\label{eq:evModel3}
\mathcal{L}_\text{evs, norm} = \| \frac{\Delta \hat{\mathbf{L}}(\Theta)}{\| \Delta \hat{\mathbf{L}}(\Theta) \|_2} - \frac{\Delta \mathbf{L}(\Theta)}{\| \Delta \mathbf{L}(\Theta) \|_2} \|_2^2.
\end{equation}

\subsection{Event Sampling}
In each optimization step, we include a large number of event pairs from different pixels and at different times. An event pairs consists of an event and its successor event in time (at the same pixel). Our method is general and can also work on event pairs $\{e_k, e_{k+j}\}$ which are not direct neighbors but further apart in time ($j>1$). In this case, we accumulate the polarities of all events occurring between the chosen events, i.e. we set the measured brightness difference in our loss function to $\Delta L_k = C \, \sum_{i=k+1}^{i=k+j} p_{i}$. 
Uniform brightness areas do not trigger events. We can include this fact in our model by also sampling from areas where no event has occurred for a time period of $T_{\text{noevs}}$. A non-existing event means that $L$ did neither increase nor decrease by more than C, hence we formulate the no-event loss as

\begin{equation} \label{eq:noevModel}
\mathcal{L}_{\text{noevs}} = \sum_{k} \mathop{relu}(|\hat{L}_{k} - \hat{L}_{k-1}| - C).
\end{equation}

We precompute no-event locations by saving their pixel coordinates $\textbf{u}_\text{noev}$ as well as the beginning $t_0(\textbf{u}_\text{noev})$ and end $t_1(\textbf{u}_\text{noev})$ of the time interval $T_{\text{noevs}}$. During training we first sample a random no-event pixel and then sample two random timestamps in the interval $[t_0(\textbf{u}_\text{noev}), t_1(\textbf{u}_\text{noev})]$. Overall, if the no-event loss is enabled, we sample two thirds of event rays and one third of no-event rays. The final loss is a weighted combination of $\mathcal{L}_\text{evs} + \lambda_\text{noevs} \, \mathcal{L}_\text{noevs}$.

\subsection{Implementation Details}
We base our code on torch-ngp~\cite{torch-ngp} employing a hash-based encoding \cite{mueller2022instant}. The training usually converges on one NVIDIA A40 in about 1 to 3 hours  (including elaborate logging and preprocessing). Larger batch sizes can improve PSNR values slightly and converge faster. Given external poses from a motion capture system, for each sampled event we interpolate the rotational component of its pose $\mathbf{T}(t_k)$ by spherical linear interpolation (slerp), and the translational component by cubic interpolation.

\section{EXPERIMENTS}
We evaluate the proposed E-NeRF on synthetic data based on ESIM~\cite{corl_RebecqGS18} as well as on real-world, public benchmarks. \\
We compare E-NeRF to several baselines: (i)~frame-based torch-ngp~\cite{torch-ngp} (ii)~frame-based Deblur-NeRF~\cite{ma2021deblur} when considering scenarios with motion blur, and (iii)~our novel baseline method consisting of frame reconstruction using E2VID~\cite{DBLP:journals/pami/RebecqRKS21} followed by torch-ngp~\cite{torch-ngp}\footnote{E2VID yields strong artefacts during fast motion, degrading its performance significantly. For this reason, to be fair to the E2VID baseline, we only feed a subset of all events into E2VID (between 1/4th and 1/16th) during high event rates, and report the best result. This upsampling is common practice and required for good results, as noted in the \href{https://github.com/uzh-rpg/e2calib\#upsampling}{official code} of \cite{DBLP:conf/cvpr/MuglikarGG021}.}. 
While RawNeRF \cite{mildenhall2021rawnerf} would be a good additional baseline it cannot be used because there was no public code at the time of writing and because it requires raw images from a camera that is static during the exposure. Following the motivation of this work, we consider scenarios that are frequent in robotics and challenging for frame-based (and event-based) vision.

\subsection{Synthetic Data}
We simulate five sequences with different motion trajectories and background textures using the event camera simulator ESIM \cite{corl_RebecqGS18}. Quantitative image metrics are computed between rendered frames and holdout test frames without motion blur. Since the absolute brightness is non-observable from the event stream (see \autoref{eq:evModel}), we perform an affine brightness transform on our predictions using the holdout test frames for all presented methods. The evaluation uses the peak signal to noise ratio (PSNR), the structural similarity (SSIM)~\cite{wang2004image}, and the Learned Perceptual Image Patch Similarity (LPIPS)~\cite{DBLP:conf/cvpr/ZhangIESW18} using AlexNet.

\begin{figure}[tbp]
     \centering
     \begin{subfigure}[t]{0.23\textwidth}
         \centering
         \includegraphics[width=\textwidth]{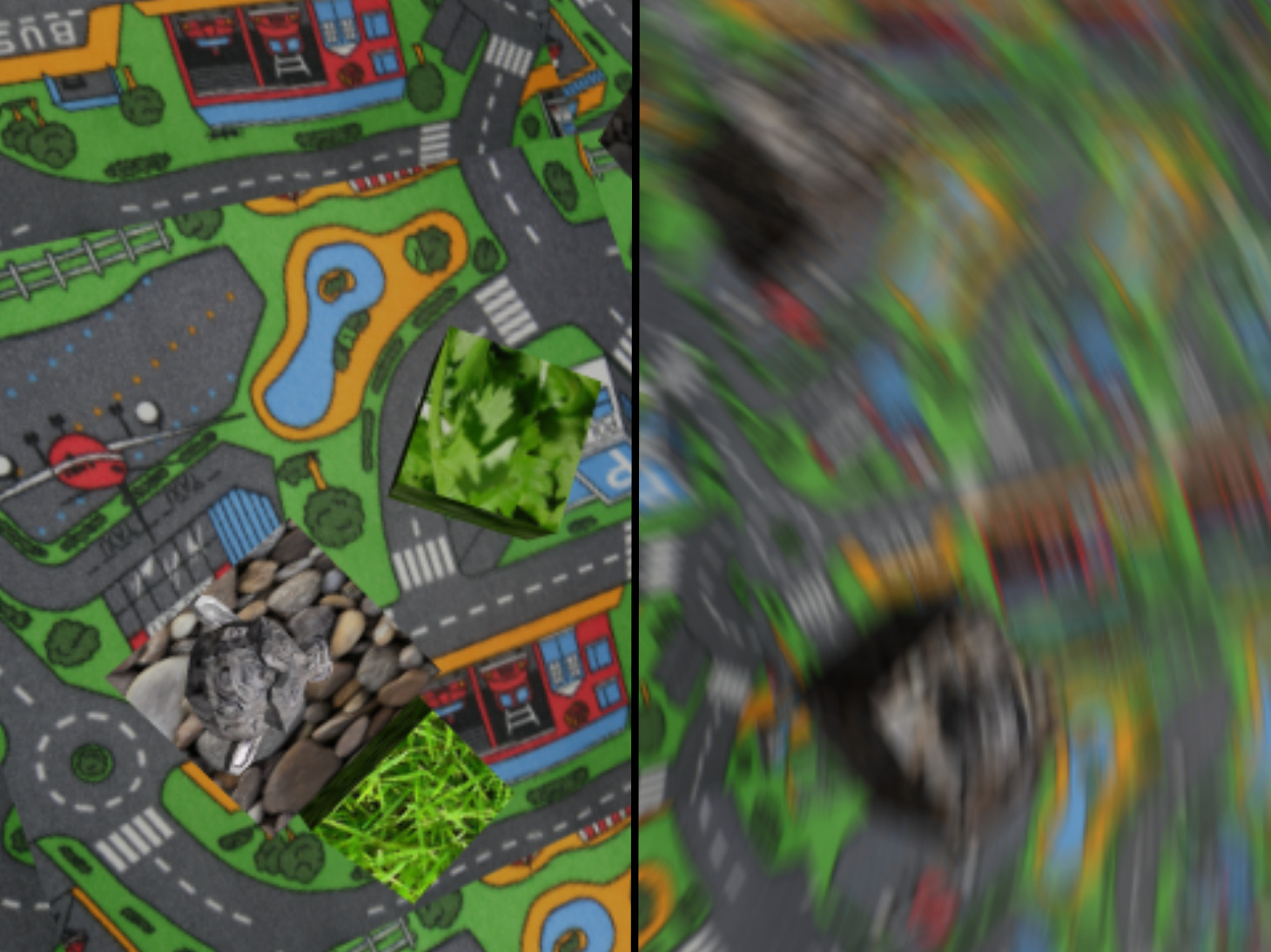}
         \vspace{-1.3\baselineskip}
         \caption{RGB holdout/train frame}
     \end{subfigure}
     \hfill
     \begin{subfigure}[t]{0.23\textwidth}
         \centering
         \includegraphics[width=\textwidth]{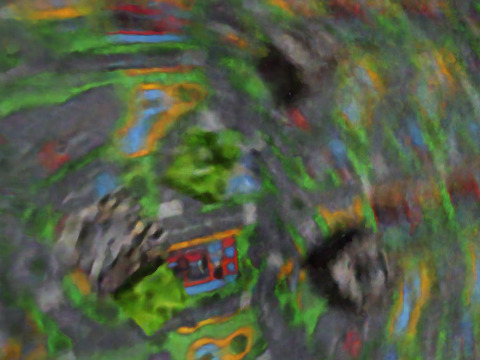}
         \vspace{-1.3\baselineskip}
         \caption{Deblur-NeRF \cite{ma2021deblur} baseline}
     \end{subfigure}\\[0.9ex]
     \begin{subfigure}[t]{0.23\textwidth}
         \centering
         \includegraphics[width=\textwidth]{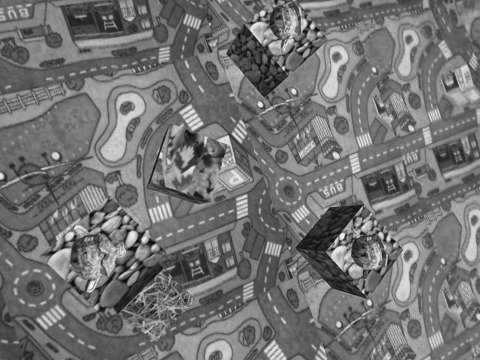}
         \vspace{-1.3\baselineskip}
         \caption{E-NeRF (event-only)}
    \end{subfigure}
         \hfill
    \begin{subfigure}[t]{0.23\textwidth}
         \centering
         \includegraphics[width=\textwidth]{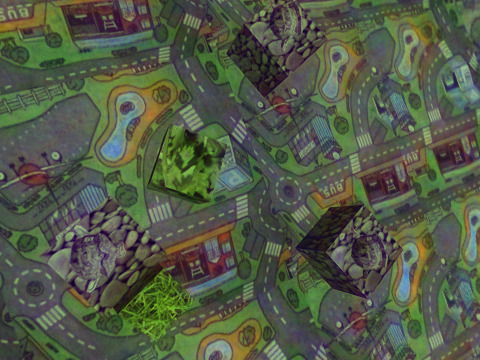}
         \vspace{-1.3\baselineskip}
         \caption{E-NeRF (events \& frames w/ blur).}
     \end{subfigure}
     \caption{Qualitative evaluation of sequence Shake Carpet 1. (a)~The input views contain strong motion blur. (b)~The frame-only baseline Deblur-NeRF \cite{ma2021deblur} can not recover sharp details. (c)~In contrast, E-NeRF recovers sharp details from event data only. (d)~When combining grayscale events and blurry color frames, our method is able to produce sharp and colored reconstructions. Due to only using grayscale events the colors are not photorealistic, but could still be useful for object recognition.}
     \vspace{-1.3\baselineskip}
     \label{fig::synt_motion_blur}
\end{figure}

\pseudoparagraph{Motion Blur}
We simulate images with motion blur in ESIM by integrating irradiance maps over the exposure time, which yields realistic, motion-dependent blur \cite[Sec.~7.4]{corl_RebecqGS18}. \autoref{tab::spiral} and \autoref{tab::shake} show that our method produces better results than the baselines.
Notably, the results are much better in comparison to combining E2VID with frame-based torch-ngp. While E2VID has the advantage of being trained on a large dataset, the proposed E-NeRF fuses information in 3D and can leverage spatial consistency without requiring any training data.
Additionally, E-NeRF shows better results than frame-based methods, which cannot overcome the strong motion blur. While Deblur-NeRF shows good results for the mild blur in \autoref{tab::shake}, it is worse than torch-ngp for the strong blur in \autoref{tab::spiral}. This result can be expected because the blur is consistent, i.e. not stemming from random motion in all directions, and the training views are not dense. This shows that carefully modelling the blur process can produce better results, but using blur-resistant event data is preferable.
Finally, the results in \autoref{fig::synt_motion_blur} show that E-NeRF is able to reconstruct sharp details, which can be beneficial for robotics applications. For example for licence plate detection it is important that the reconstruction is sharp and the characters are legible, while an overall photorealistic reconstruction is less important.

\input{tables/shake}

\pseudoparagraph{Combining Events and Frames}
We combine events and blurry frames by a simple weighted combination 
\begin{equation}
    \mathcal{L} = \mathcal{L}_\text{evs} + \lambda_\text{rgb} \, \mathcal{L}_\text{rgb},
\end{equation}
\noindent where $\mathcal{L}_\text{rgb}$ is the original RGB rendering NeRF loss  \cite{mildenhall2020nerf}. Using events yields sharp edges but uniform areas show fog-like artefacts due to the discrete nature of events. Additionally, the color for one shaded area might be slightly off because it is never directly measured but inferred from derivative-like data. Frame data can help to correct both problems.
In \autoref{fig::synt_motion_blur} we combine grayscale events and blurry color frames into a sharp and colored reconstruction. We do not require a color event camera but simply map the predicted color values of the event rays to grayscale. Note that most commercially available event cameras are grayscale.
Moreover, combining events and frames opens up potential low-power reconstructions, where frames are triggered at low-frequency and the event camera captures necessary data in the blind time.

\pseudoparagraph{Ablation Study}\label{par::ablation}
We study our method by changing single components and report the average image metrics over all five simulated sequences. 
Changing the loss function from \autoref{eq:evModel2} to the normalized loss in \autoref{eq:evModel3} results in a minor drop of performance (see line~(i) in \autoref{tab::ablation}). This is expected, since the normalized loss is an approximation of the ideal physical model. However, the drop in performance is not very large, and we find the normalized loss function to work better on some of the real-world data (e.g. \autoref{fig:eds00}).

Our default sampling strategy is to sample the closest event neighbors in time, which results in an average event pair distance of around 1ms. We investigate a different sampling with event pairs being farther apart in time. To do this, we first divide the event stream into windows of 60ms. For each sampled event, we pick the last event at that pixel in the 60ms window as neighbor, resulting in an average event pair distance of approximately 30ms.
We show in line~(ii) of table \autoref{tab::ablation} that by changing the sampling strategy as described, we receive a minor drop in performance. We believe that this drop might be due to sampling some event rays repeatedly (at the end of the 60ms window).
Since this strategy results in unstable training on Shake Moon 1 and Shake Moon 2, we only accumulate up to five events on those sequences.

In line~(iii) of table \autoref{tab::ablation} we show the influence of adding the no-event loss \autoref{eq:noevModel} with a weight of 0.1 to the event loss function. We search for no-events in the time interval $T_{\text{noevs}} = 25$ms and allow up to one fourth of all event rays in the batch to be no-events. This configuration also yields a minor drop in performance. However, we find that the no-event loss helps to remove artefacts in uniformly colored areas, see \autoref{fig::noEvsReal}. Moreover, using the no-event loss as well as the event accumulation, can help to improve the visual quality on real data, e.g. in \autoref{fig:eds11} and \autoref{fig:eds00}.

\input{tables/ablation}

\subsection{Real Data}
\begin{figure*}
	\centering
	\footnotesize
    \begin{tabular}{c c c c c}
    \includegraphics[width=0.17\linewidth]{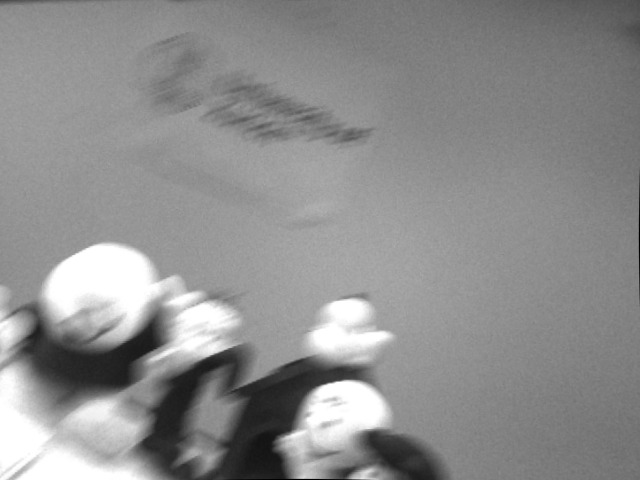}
    & \includegraphics[width=0.17\linewidth]{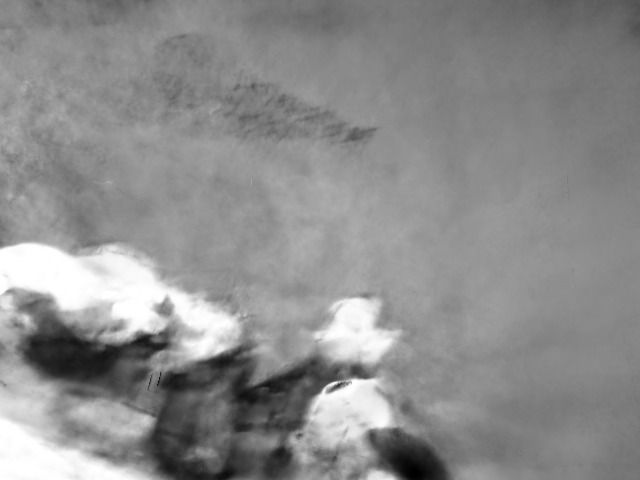}
    & \includegraphics[width=0.17\linewidth]{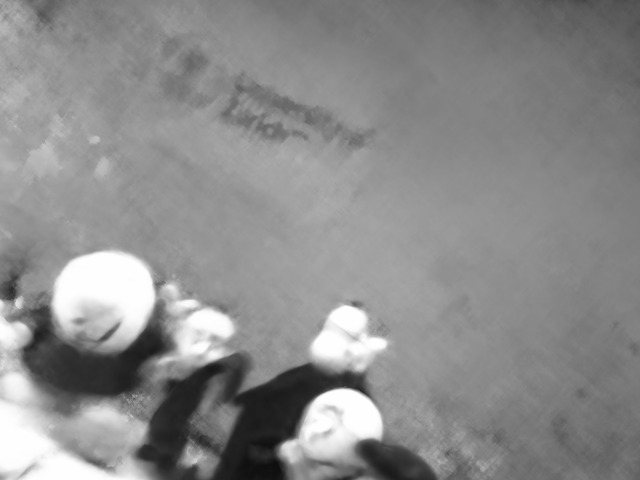}
    & \includegraphics[width=0.17\linewidth]{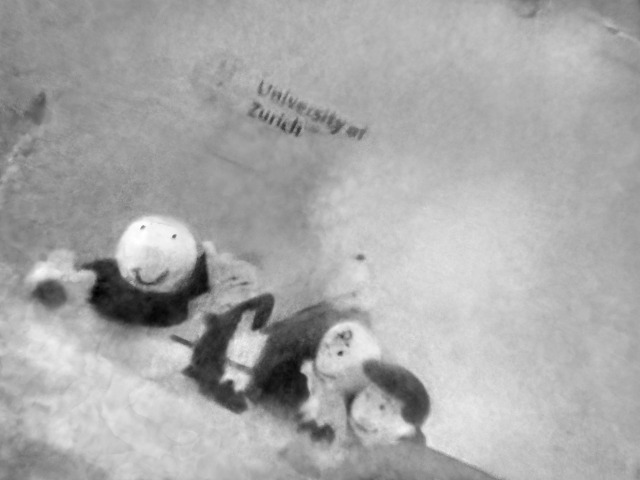}
    & \includegraphics[width=0.17\linewidth]{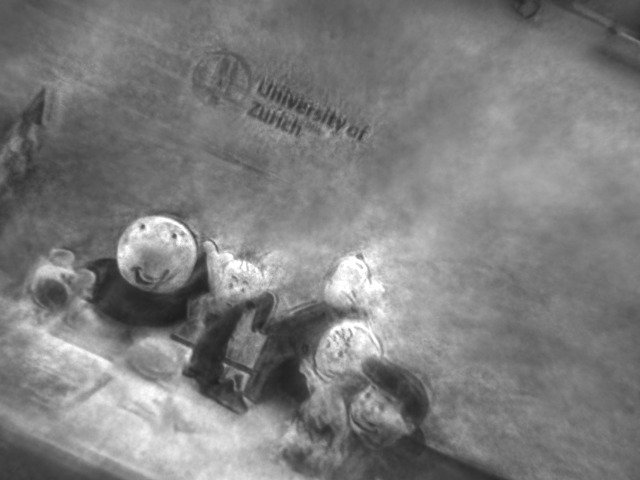}\\[0.2ex]
    \includegraphics[width=0.17\linewidth]{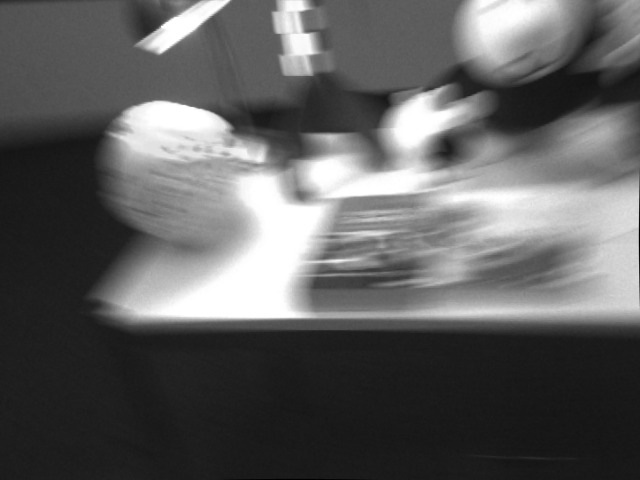}
    & \includegraphics[width=0.17\linewidth]{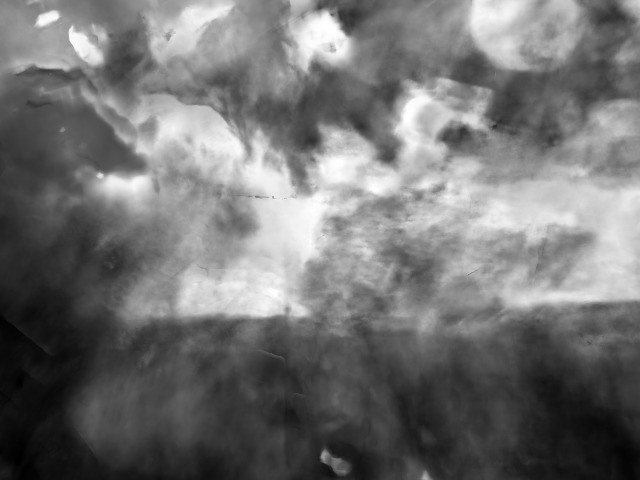}
    & \includegraphics[width=0.17\linewidth]{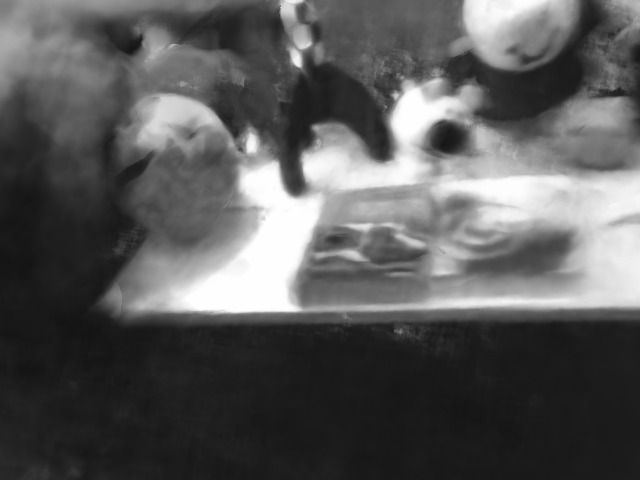}
    & \includegraphics[width=0.17\linewidth]{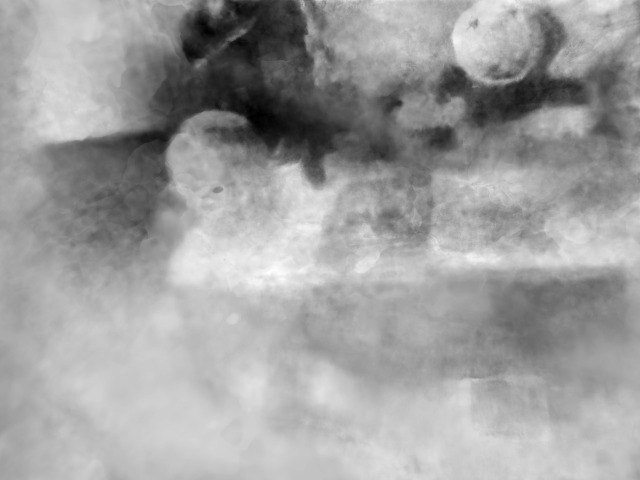}
    & \includegraphics[width=0.17\linewidth]{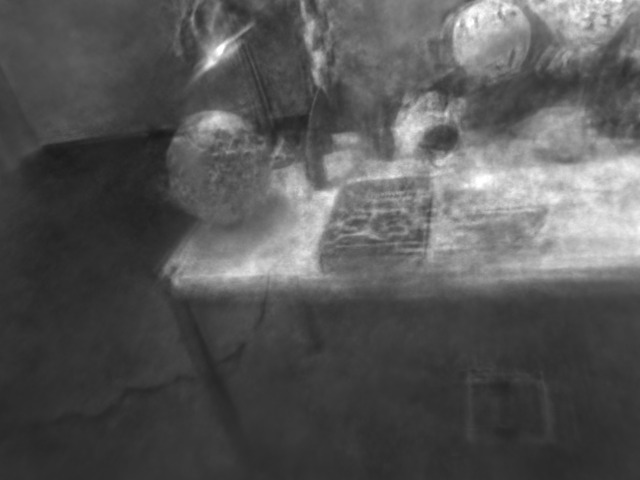}\\[0.2ex]
    \includegraphics[width=0.17\linewidth]{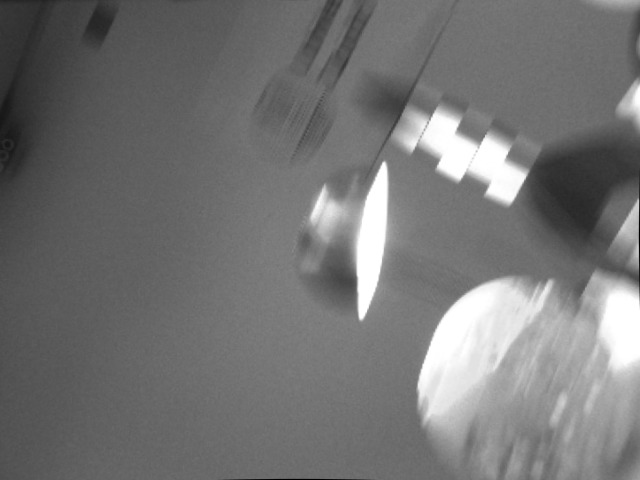}
    & \includegraphics[width=0.17\linewidth]{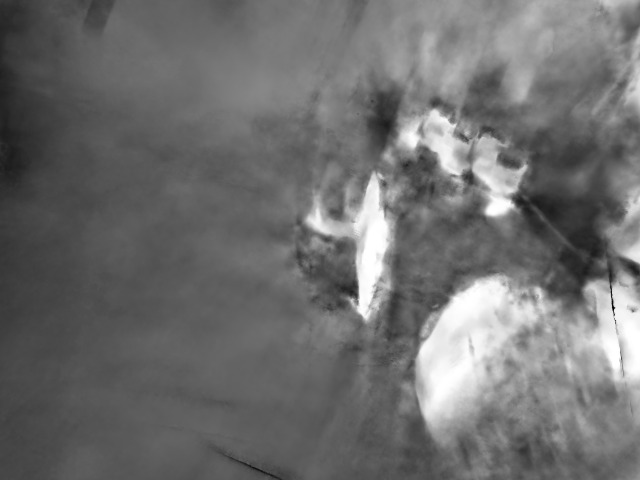}
    & \includegraphics[width=0.17\linewidth]{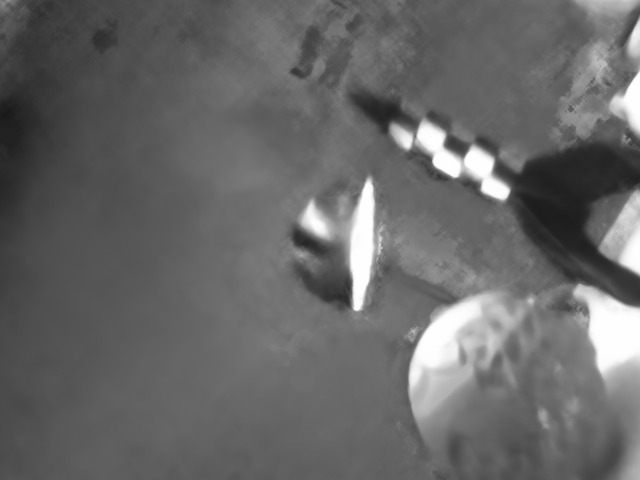}
    & \includegraphics[width=0.17\linewidth]{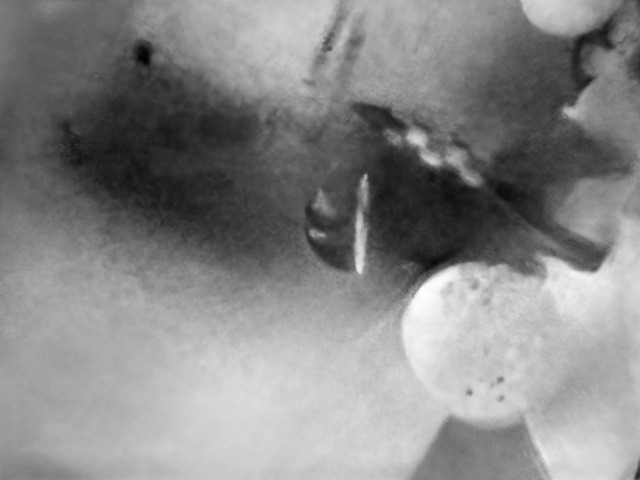}    
    & \includegraphics[width=0.17\linewidth]{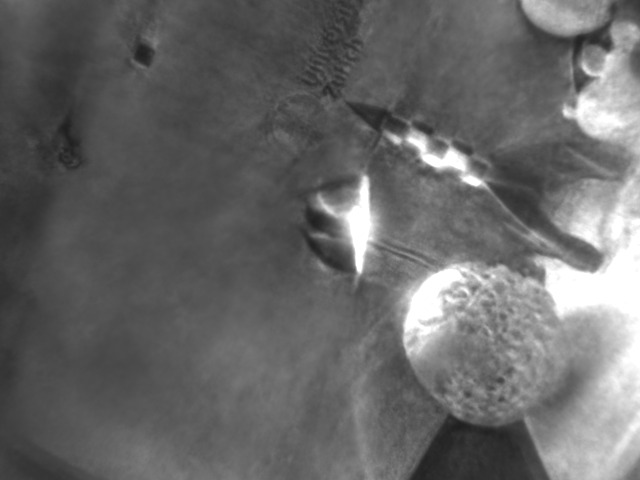}\\[0.2ex]
    \includegraphics[width=0.17\linewidth]{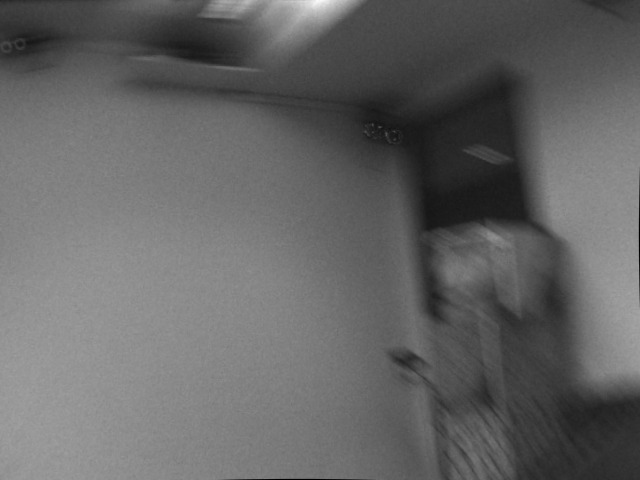}
    & \includegraphics[width=0.17\linewidth]{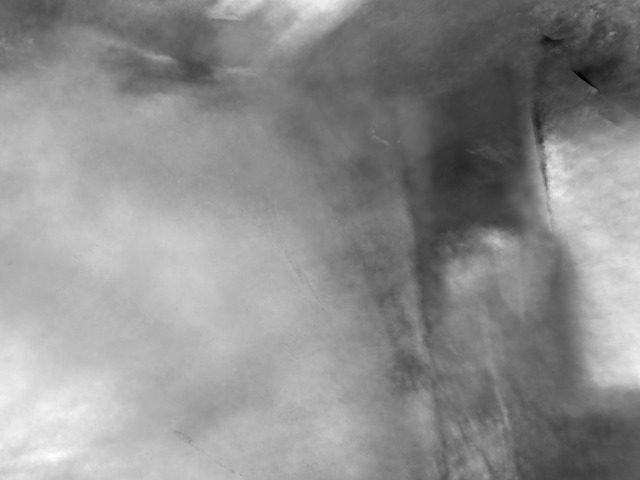}
    & \includegraphics[width=0.17\linewidth]{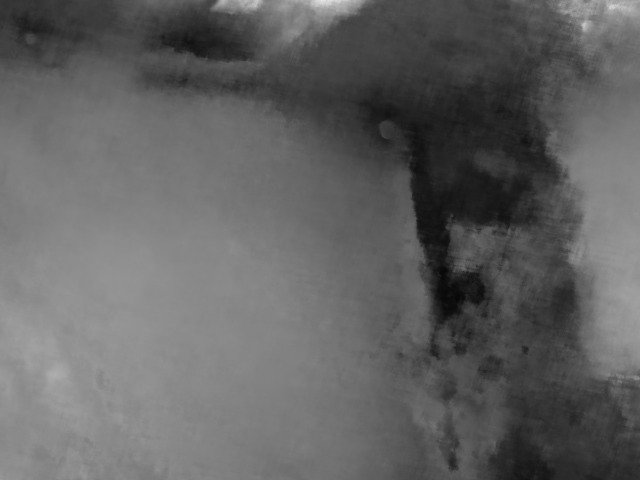}
    & \includegraphics[width=0.17\linewidth]{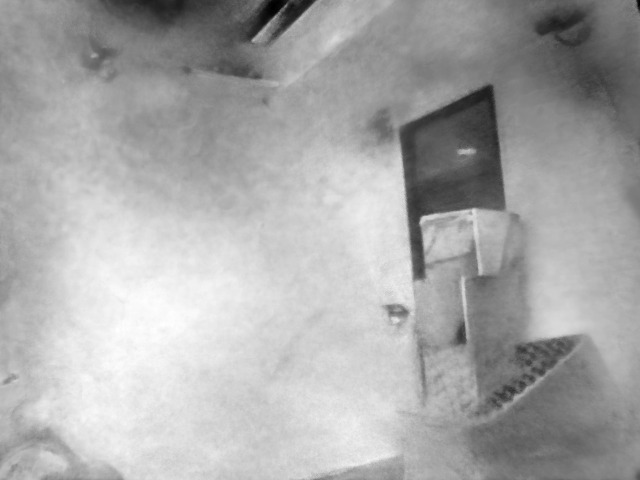}
    & \includegraphics[width=0.17\linewidth]{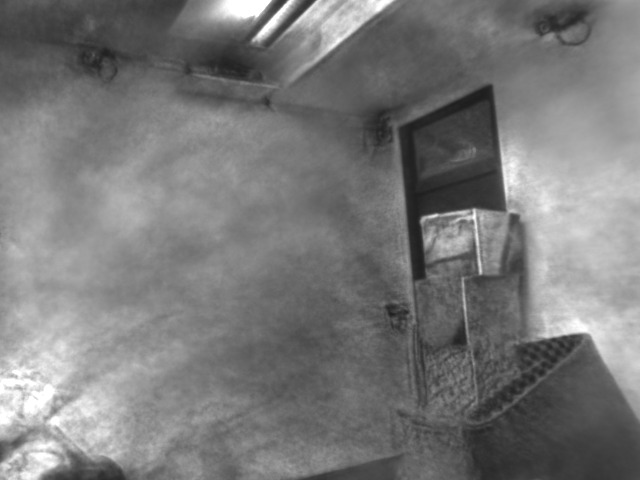}\\[0.2ex]
    (a) holdout frame & (b) torch-ngp~\cite{torch-ngp} & (c) Deblur-NeRF~\cite{ma2021deblur} & (d) E2VID~\cite{DBLP:journals/pami/RebecqRKS21}+torch-ngp & (e) E-NeRF (event-only)\\
    \end{tabular}
\caption{Qualitative evaluation on EDS-11 during high speed motion. Holdout frames (a) are not used during optimization. Due to very strong motion blur, the NeRF baseline (b) can not recover proper geometry. The Deblur-NeRF baseline (c) can reconstruct the scene, but shows artefacts on the wall and is missing details such as the eyes of the toy figure or the university logo. The E2VID+torch-ngp baseline (d) shows good performance on uniformly colored areas, while not being able to reconstruct detailed geometry and texture. Our method (e) can better reconstruct sharp details, such as the lettering, the book cover, the grid structure of the ball, as well as the texture of the boxes in the last row. The E-NeRF result is obtained by picking a fixed threshold of $C=0.2$, and by adding $\mathcal{L}_{\text{noevs}}$ \autoref{eq:noevModel} with weight one to the loss ($T_{\text{noevs}} = 30$ms), which improves visual quality on the white background. E2VID uses upsampling of factor 16. We use a batch size of 30096 event pairs.}.
\vspace{-1.0\baselineskip}
\label{fig:eds11}
\end{figure*}

\begin{figure*}
	\centering
	\footnotesize
    \begin{tabular}{c c c c c}
    \includegraphics[width=0.17\linewidth]{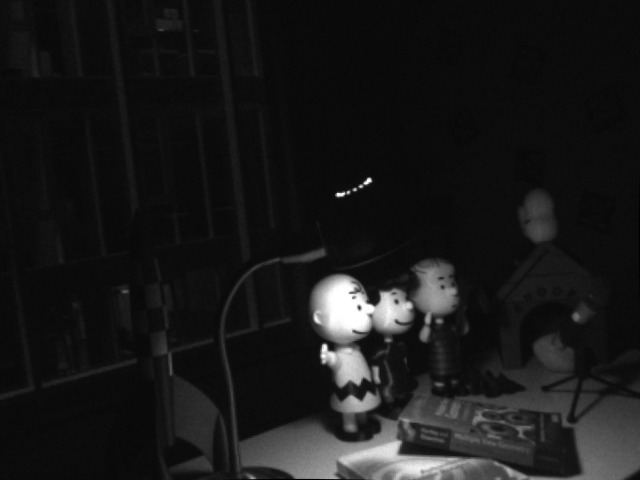}
    & \includegraphics[width=0.17\linewidth]{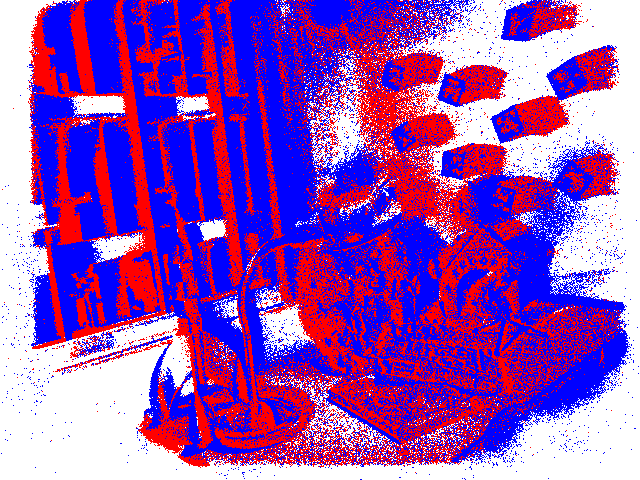}
    & \includegraphics[width=0.17\linewidth]{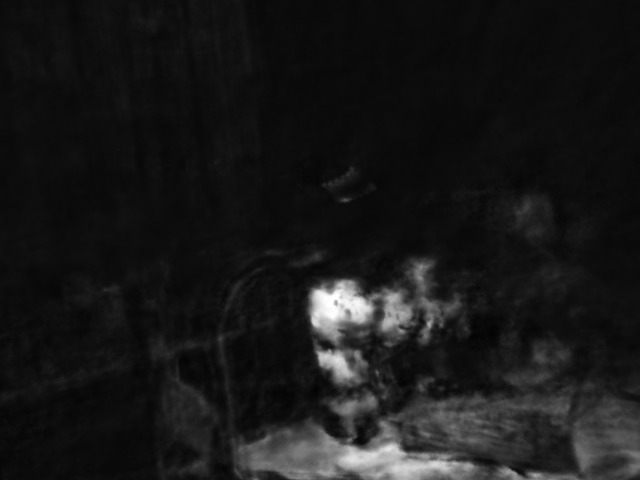}
    & \includegraphics[width=0.17\linewidth]{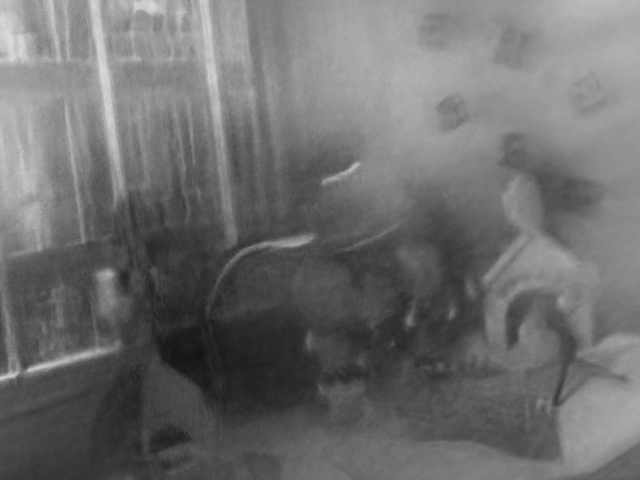}
    & \includegraphics[width=0.17\linewidth]{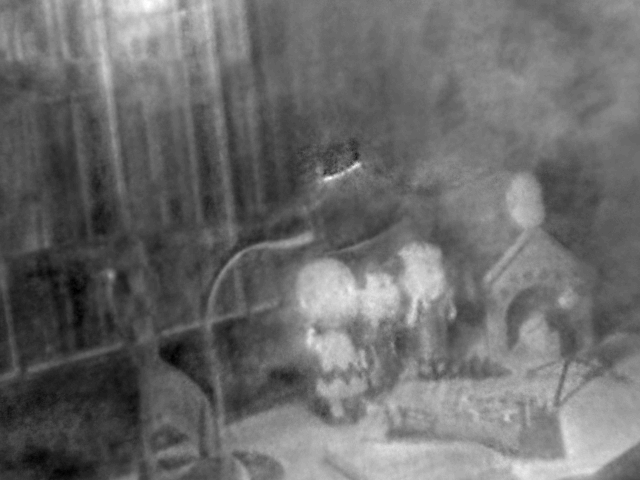}\\[0.2ex]
    \includegraphics[width=0.17\linewidth]{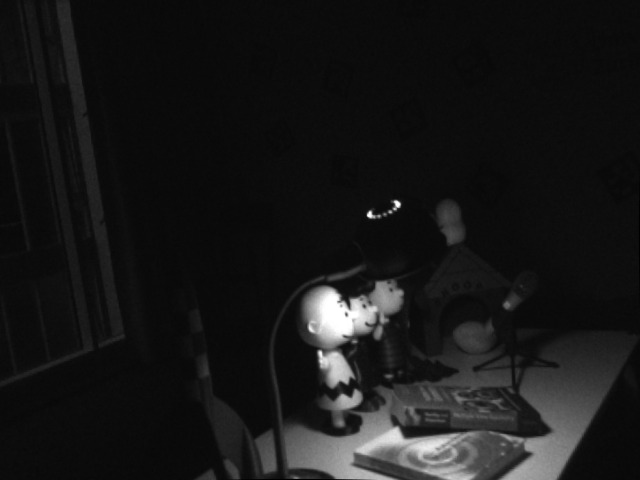}
    & \includegraphics[width=0.17\linewidth]{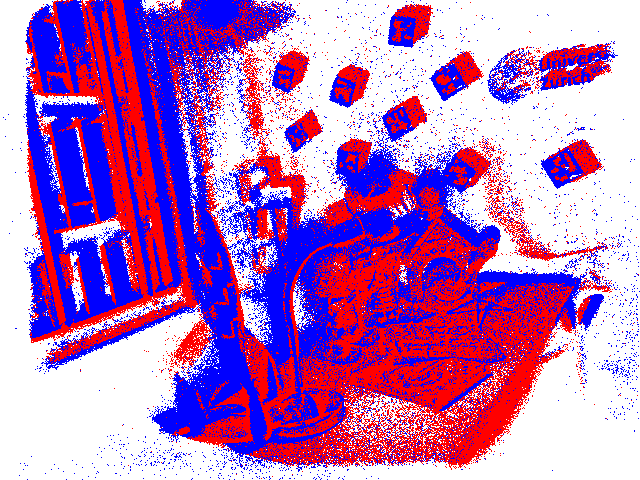}
    & \includegraphics[width=0.17\linewidth]{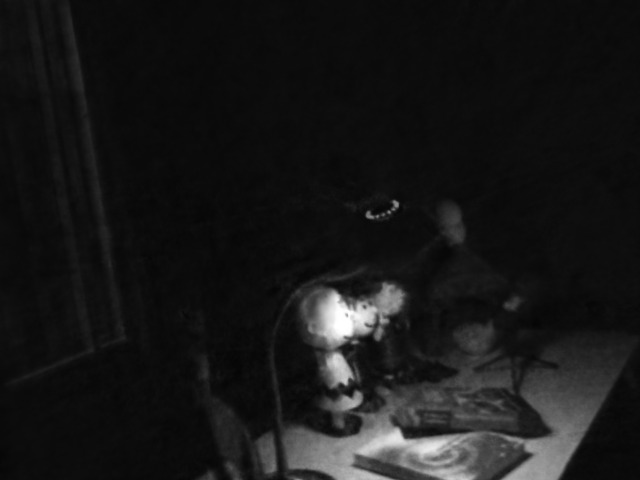}
    & \includegraphics[width=0.17\linewidth]{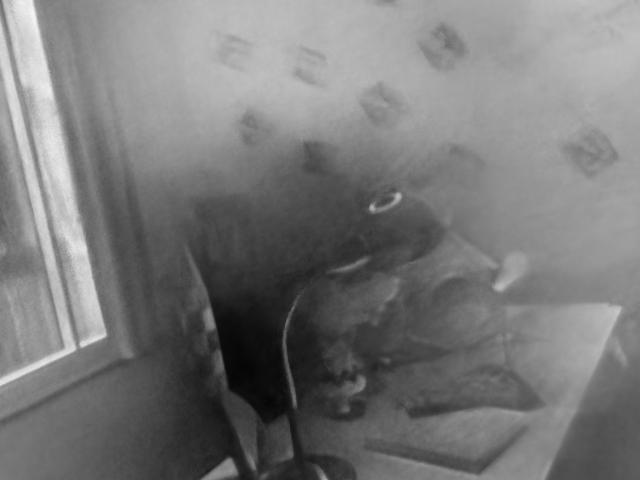}
    & \includegraphics[width=0.17\linewidth]{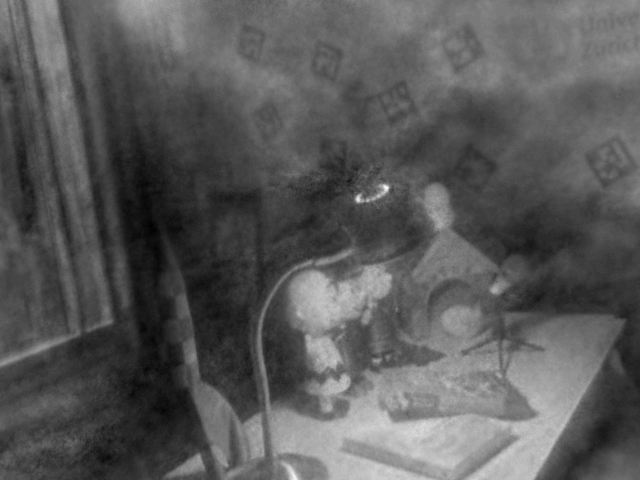}\\[0.2ex]
    \includegraphics[width=0.17\linewidth]{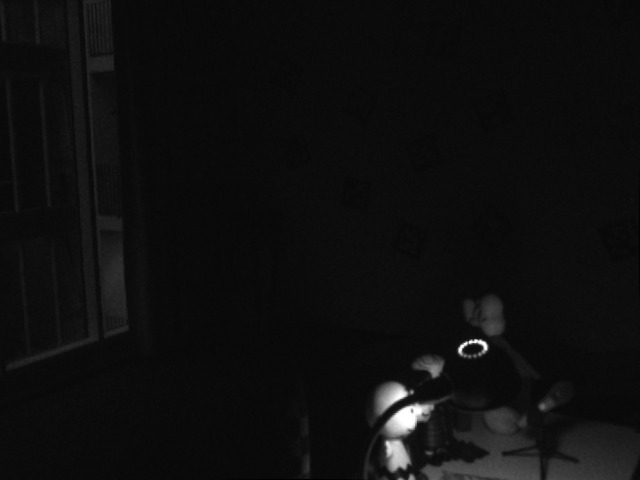}
    & \includegraphics[width=0.17\linewidth]{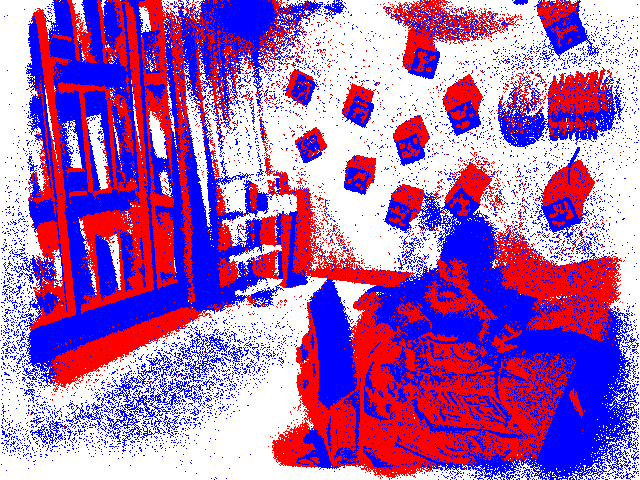}
    & \includegraphics[width=0.17\linewidth]{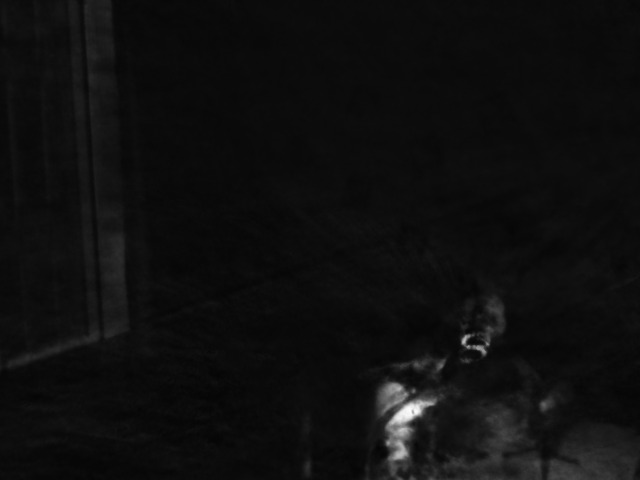}
    & \includegraphics[width=0.17\linewidth]{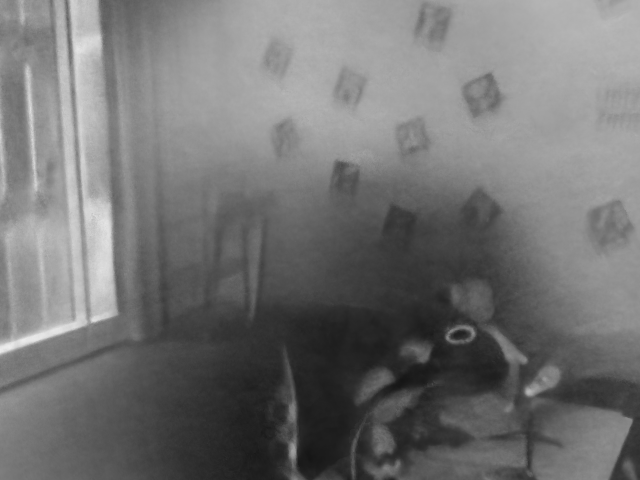}
    & \includegraphics[width=0.17\linewidth]{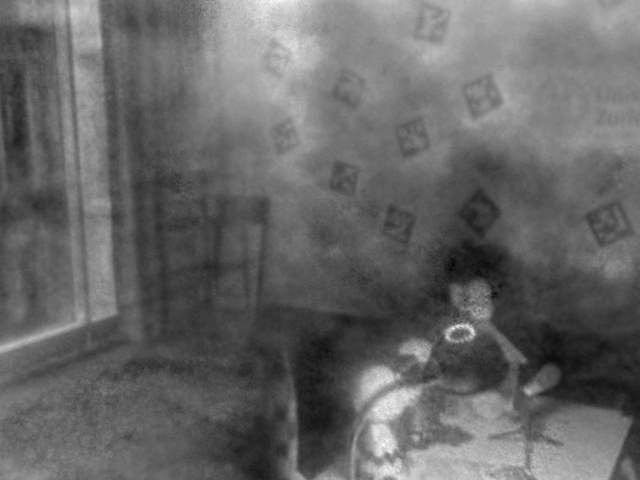}\\[0.2ex]
    (a) holdout frame & (b) Events & (c) torch-ngp~\cite{torch-ngp} & (d) E2VID~\cite{DBLP:journals/pami/RebecqRKS21}+torch-ngp& (e) E-NeRF (event-only)\\
    \end{tabular}
\caption{Qualitative evaluation on EDS-00. We show the holdout frames (a), as well as the closest input events (b) to E2VID (d) and E-NeRF (e). Notice how the NeRF-baseline (c) shows geometry artefacts for some novel viewpoints and can not reconstruct the wall at the back of the room due to limited dynamic range. The E2VID+torch-ngp (d) baseline can recover details in the dark regions, but shows overly dark reconstructions of the illuminated figures. E-NeRF (e) can reconstruct sharp details in the foreground, such as the book cover and detailed clothing. E-NeRF can even reconstruct the writing on the background wall in the top-right corner of the third and fourth row. The E-NeRF result is obtained by using the normalized loss \autoref{eq:evModel3}, adding $\mathcal{L}_{\text{noevs}}$ \autoref{eq:noevModel} with weight one to the loss ($T_{\text{noevs}} = 30$ms), and sampling from event windows of up to 130ms. We perform a simple contrast normalization on the E-NeRF renderings for better visual clarity. E2VID uses upsampling of factor of 4. We use a batch size of 30096 event pairs.}.
\label{fig:eds00}
\end{figure*}

\pseudoparagraph{EDS}
We evaluate our method on the public dataset accompanying EDS \cite{hidalgo2022event}, which uses a beamsplitter to capture aligned frames and events from the same viewpoint\footnote{Since the event and frame camera have different intrinsics parameters (which we use for training and evaluation), the field of view differs slightly in the shown renderings \autoref{fig:eds00} and \autoref{fig:eds11}.}. We would like to note that for real data, there is no quantitative evaluation possible, since we pick sequences where the frames are severely degraded by motion blur or their low dynamic range. We use sequence 00 to evaluate our method in the dark, as well as sequence 11 to evaluate high speed motion. The selected scenarios are highly challenging for both camera modalities. The darkness in sequence 00 causes high noise levels in both cameras \cite{v2e}. In sequence 11 the camera is moving at very high speed, which induces strong motion blur in the frames and a very high event rate which can lead to single events being dropped \cite{feedbackControlEvsDelbruck}. For the high-dynamic-range sequence 00 we show the results in \autoref{fig:eds00}. While frame-based torch-ngp can give decent reconstructions in the bright areas of the scene, it struggles with the dark parts. Both event-based methods, E2VID+torch-ngp and E-NeRF, show more details. The proposed E-NeRF shows slightly fewer artefacts than E2VID+torch-ngp especially on foreground objects.
For the high-speed sequence 11 we show the results in \autoref{fig:eds11}. Deblur-NeRF provides better reconstructions than frame-based torch-ngp, but both methods fail to reconstruct sharp details like lettering in the background. E2VID+torch-ngp and E-NeRF can reconstruct these details and furthermore the proposed E-NeRF reconstructs the geometry of the table, which E2VID+torch-ngp fails to achieve.
Overall, event-based methods are superior for the presented real-world low light and high-speed scenarios. Furthermore, E-NeRF delivers reconstructions with better high-frequency details and better geometry in comparison to E2VID+torch-ngp. No method achieves photorealistic quality, however, this is not necessary for many robotics applications and might be infeasible due to the difficulty of the tackled scenario.

\global\long\def\widthplot{0.16\linewidth} %

\begin{figure*}
	\centering
	\footnotesize
    \begin{tabular}{c c c c c}
    \rotatebox{90}{\;\;(i)~60 train views}
    \includegraphics[width=3cm,height=2.3cm]{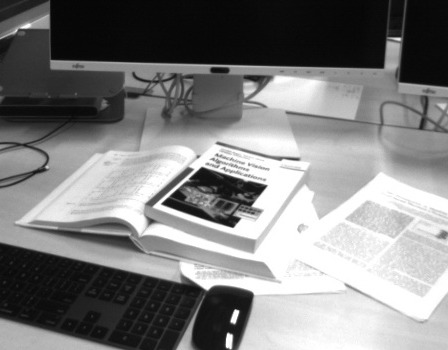}
    & \includegraphics[width=3cm,height=2.3cm]{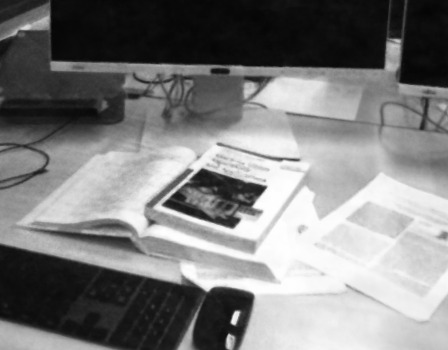}
    & \includegraphics[width=3cm,height=2.3cm]{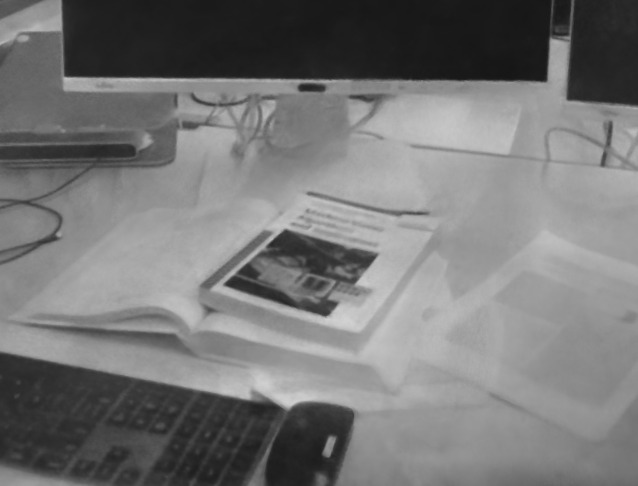}
    & \includegraphics[width=3cm,height=2.3cm]{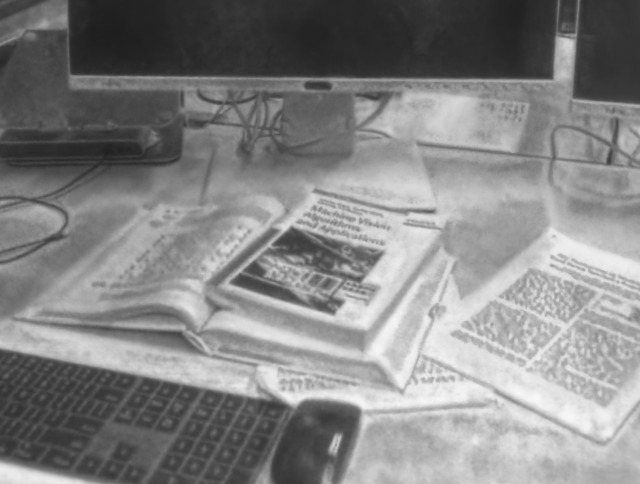}
    & \includegraphics[width=3cm,height=2.3cm]{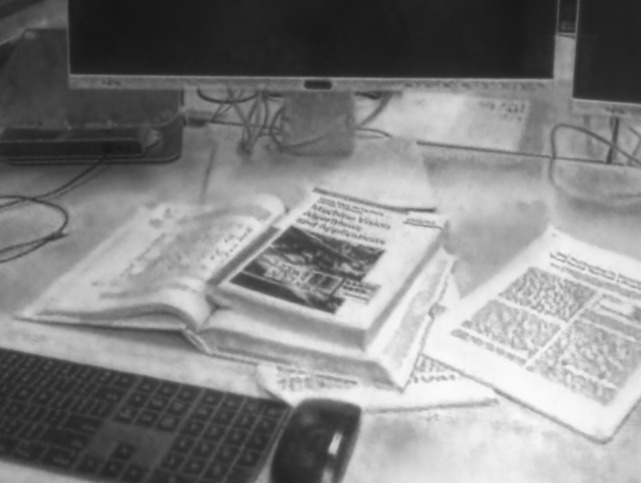}\\[0.2ex]
    \rotatebox{90}{\;\;\;(i)~6 train views}
    \includegraphics[width=3cm,height=2.3cm]{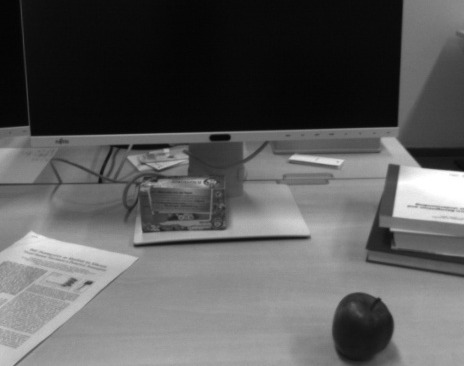}
    & \includegraphics[width=3cm,height=2.3cm]{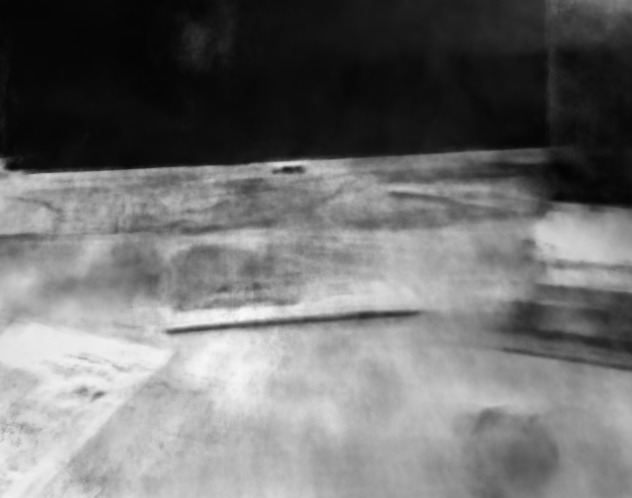}
    & \includegraphics[width=3cm,height=2.3cm]{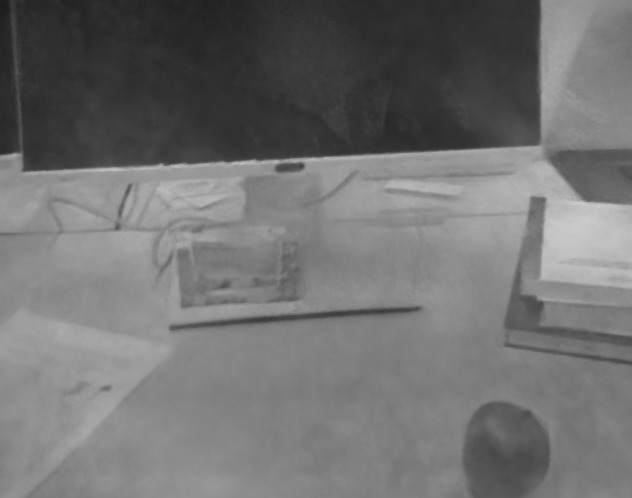}
    & \includegraphics[width=3cm,height=2.3cm]{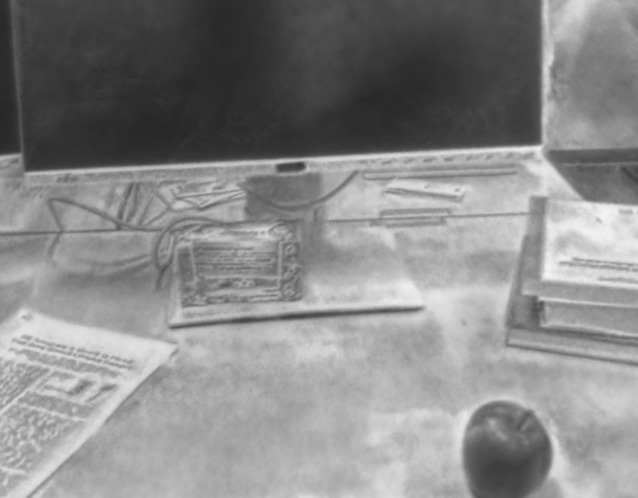}
    & \includegraphics[width=3cm,height=2.3cm]{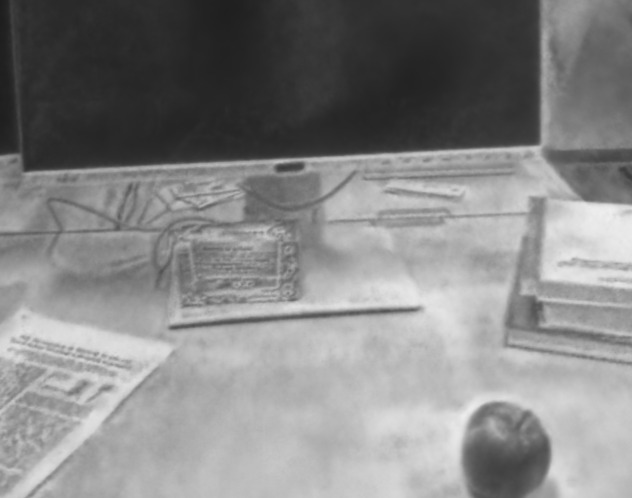}\\[0.2ex]
    (a)~holdout frame & (b)~torch-ngp~\cite{torch-ngp} & (c)~E2VID~\cite{DBLP:journals/pami/RebecqRKS21}+torch-ngp & (d)~E-NeRF (event-only) & (e)~E-NeRF (events\,+\,frames)\\
    \end{tabular}
\caption{Qualitative evaluation on TUM-VIE mocap-desk2. We crop the renderings of all methods to highlight the same part of the scene. (i)~Due to strong specularities at the table and mild motion blur, the torch-ngp baseline with 60 train views~(b) is not fully sharp and shows artefacts around the book cover and text. E2VID+torch-ngp (c) can recover more details on the dark keyboard but is not fully sharp and fails to recover text. E-NeRF (d) can reconstruct high-frequency details such as the outline of the book cover and fine-grained text on the papers. Combining events and frames (e) removes artefacts on uniformly colored areas on the table and on white spaces of the book cover and papers, while preserving high frequency details. (ii)~By using only six train views, torch-ngp is not able to reconstruct proper geometry. E2VID+torch-ngp gives decent reconstructions on the table but fails to capture fine details such as the monitor buttons and text, which E-NeRF can reconstruct properly. Combining the six train views with events (e) improves visual artefacts of E-NeRF in uniform brightness areas, e.g. on the monitor and table.
The E-NeRF result is obtained by picking a fixed threshold of $C=0.2$. E2VID uses upsampling of factor of 4. We use a batch size of 20096 event pairs.}.
\label{fig::tumvie}
\end{figure*}

\pseudoparagraph{TUMVIE}
We evaluate our method on the sequence mocap-desk2. The camera follows a forward facing, circular path inducing mild motion blur in the frames. The results in \autoref{fig::tumvie} show that E-NeRF can reconstruct fine details on the book cover and fine-grained text on the paper as well as on the opened book page which is not possible with frame-based torch-ngp, although the dataset features a high-quality uEye UI-3241LE-M-GL frame camera. We show that by combing events and frames, it is possible to improve the reconstruction in uniformly colored areas, e.g. on the table and on the dark keyboard. 
We empirically notice that the visual quality of E-NeRF is better on TUMVIE than on EDS. This might be due to the novel event camera model used in TUMVIE (Prophesee Gen4HD compared to Gen3 in EDS), as well as better illumination. In the second row of \autoref{fig::tumvie} we show that torch-ngp can not reconstruct correct geometry when we decrease the number of input views to six. 
E2VID+torch-ngp can reconstruct uniform color areas on the mocap-desk2 sequence but it does not recover the text. To qualitatively study the influence of the no-event loss $\mathcal{L}_{\text{noevs}}$ (\autoref{eq:noevModel}) on our method, we perform a hyperparameter sweep over $\mathcal{\lambda}_{\text{noevs}}$ in \autoref{fig::noEvsReal} on the scene mocap-1d-trans. The camera motion is slow and the illumination is good in this sequence. It can be noticed that uniform brightness areas, which do not trigger events, appear more smooth as $\mathcal{\lambda}_{\text{noevs}}$ increases.

\input{tables/noEvsReal.tex}

\pseudoparagraph{Discussion}
For challenging real-world scenarios in low-light conditions and with fast camera motion, E-NeRF clearly outperforms frame-based torch-ngp. Additionally, E-NeRF often shows clearer details and fewer artifacts than E2VID+torch-ngp; however, the gap is smaller than for synthetic data. The major reason is that the event generation model of \autoref{eq:evModel} is only an approximation and less accurate for these challenging scenarios. Switching to a different event camera model generally amounts to changing the loss function, which makes E-NeRF an excellent vehicle for studying event camera models for 3D reconstruction. Ultimately, a model that takes into account all spatial and temporal dependencies might be hard to specify and should be learned from data. We consider this a fruitful direction for future research that could lead to real-world E-NeRF reconstructions that are on par with reconstructions generated from synthetic data.

\section{Conclusion}
We present E-NeRF, the first method that reconstructs a neural radiance field from a fast-moving event camera. We specifically focus on scenarios that are common in robotics, such as strong motion blur or non-ideal illumination in the dark. We show on synthetic and real-world data that E-NeRF outperforms multiple baselines, and we ablate our design choices. In particular, we show that E-NeRF is able to better reconstruct high-frequency details such as text.
We show that, if only a few input views are provided, an additional event stream helps to estimate NeRFs. Our method can even combine color frames with grayscale events to obtain a sharp, colored reconstruction that utilizes the strengths of both sensors.  We see promising future work in the direction of devising better sensor models for event cameras (under challenging capturing conditions) and incorporating learning-based priors.

\ifCLASSOPTIONcaptionsoff
  \newpage
\fi

\clearpage


\section*{Supplementary material}

# Supplementary Material
# E-NeRF: Neural Radiance Fields from a Moving Event Camera

Simon Klenk[1], Lukas Koestler[1], Davide Scaramuzza[2], Daniel Cremers[1]

## I. SUPPLEMENTARY MATERIAL

### A. Details on no-event loss $\mathcal{L}_{noevs}$

A non-existing event is defined if no event is triggered in a predefined time interval $T_{\text{noevs}}$ at a pixel $\mathbf{u}_{\text{noev}}$. For computational reasons, we precompute the no-event locations by saving their pixel coordinates $\mathbf{u}_{\text{noev}}$ as well as the beginning $t_0(\mathbf{u}_{\text{noev}})$ and end $t_1(\mathbf{u}_{\text{noev}})$ of the time interval in which no event has been detected. During training, in addition to sampling events, we sample from the no-event locations. We do this by sampling two random timestamps in the interval $[t_0(\mathbf{u}_{\text{noev}}), t_1(\mathbf{u}_{\text{noev}})]$, and subsequently compute the corresponding rays $\mathbf{r}_{\text{noev0}}$ and $\mathbf{r}_{\text{noev1}}$. Overall, if the no-event loss is enabled, we sample two thirds of event rays and one third of no-event rays. The final loss is a weighted combination

$$\mathcal{L} = \mathcal{L}_{\text{evs}} + \lambda_{\text{noevs}} \mathcal{L}_{\text{noevs}}. \tag{1}$$

As can be seen in Fig. 1, the no-event sampling is noticeable inside large, uniformly colored areas, e.g. in the landing area of the plane or on the roads of the carpet. The overall reconstruction gets more smooth with increasing values of $\lambda_{\text{noevs}}$. In Tab. I it can be seen that the SSIM and LPIPS improve with negative sampling (for $\lambda_{\text{noevs}} = 1$ and $\lambda_{\text{noevs}} = 10$), whereas the PSNR value drops slightly.

| $\lambda_{\text{noevs}}$ | PSNR↑ | SSIM↑ | LPIPS↓ |
|---|---|---|---|
| 0 | 26.64 | 0.89 | 0.027 |
| 10 | 28.11 | 0.92 | 0.020 |
| 100 | 26.72 | 0.90 | 0.027 |
| 1000 | 22.43 | 0.82 | 0.065 |

TABLE I: PSNR values for different weights $\lambda_{\text{noevs}}$ of the no-event loss, on sequence Shake Carpet 1.

Please note that most of the other presented examples do not contain large uniform brightness areas, and hence we do not see a large effect of the no-event loss in the ablation study.

### B. Details on combining frames and events

We propose a very simple, weighted combination of RGB and event data and show that this works well, even if the RGB views contain heavy motion blur and are sparse (only 20 views are used in Fig. 2 in our main paper). The final loss function for combining frames and events is

[1] Computer Vision Group, Technical University of Munich, Germany.
[2] Robotics and Perception Group, University of Zurich, Switzerland.

$$\mathcal{L} = \mathcal{L}_{\text{evs}} + \lambda_{\text{rgb}} \mathcal{L}_{\text{rgb}}, \tag{2}$$

where $\mathcal{L}_{\text{rgb}}$ is the vanilla NeRF loss proposed in [1].

In the main paper, we set $\lambda_{\text{rgb}} = 1$ for simplicity. In Fig. 2, we set $\lambda_{\text{rgb}} = (0.01, 0.1, 1, 10)$ and present the resulting reconstructions. It can be noticed that for large values of $\lambda_{\text{rgb}}$ the reconstructions shows more motion blur artefacts. For small values of $\lambda_{\text{rgb}}$ the reconstructions contains less artefacts due to motion blur, but the color values increasingly loose realism, because the event camera only captures grayscale information.

### C. Additional results for slow camera motion

In contrast to the real data examples from the main paper, we provide additional results of our method on a scenario where traditional NeRF is not at a severe disadvantage. We use the scene *mocap-1d-trans* of TUM-VIE [2], where the camera is moving slowly and the illumination is good, using 90 training views for the torch-ngp baseline. We present the reconstructions in Tab. II. To be equally fair to E-NeRF and the torch-ngp baseline, we render validation views for E-NeRF and torch-ngp from their corresponding training camera view in Tab. II. To be specific, we train and validate E-NeRF in the left event camera, and we train and validate the torch-ngp baseline in the left grayscale camera.

In Tab. II we present the quantitative evaluation of *mocap-1d-trans*. Fig. 3 shows that both methods produce reasonable reconstruction qualities. However, please note that we evaluate both methods in the left grayscale camera´s viewpoint, hence E-NeRF is at a severe disadvantage due to the completely different viewpoint compared to the training views. Furthermore, small extrinsic calibration errors do also reflect in these pixel-wise metrics in E-NeRF, but not in the torch-ngp baseline which uses the same viewpoints for training and evaluation. There exists a baseline of around 4.49 centimeters between left event camera and left grayscale camera, see Fig. 2 in [2]. Since the event camera has a significantly smaller field of view, we evaluate the image quality metrics for E-NeRF and torch-ngp inside a bounding box of dimension $H \times W = 200 \times 800$ (upper left corner at $(x = 100, y = 400)$ and lower right corner at $(x = 900, y = 600)$).

### D. Runtime Analysis

In Tab. III we analyse the runtime of our program. The first row shows the runtime in seconds per training step, the rows below show the fractional runtime in percent. It can be noticed, that both frame-based torch-ngp and E-NeRF require

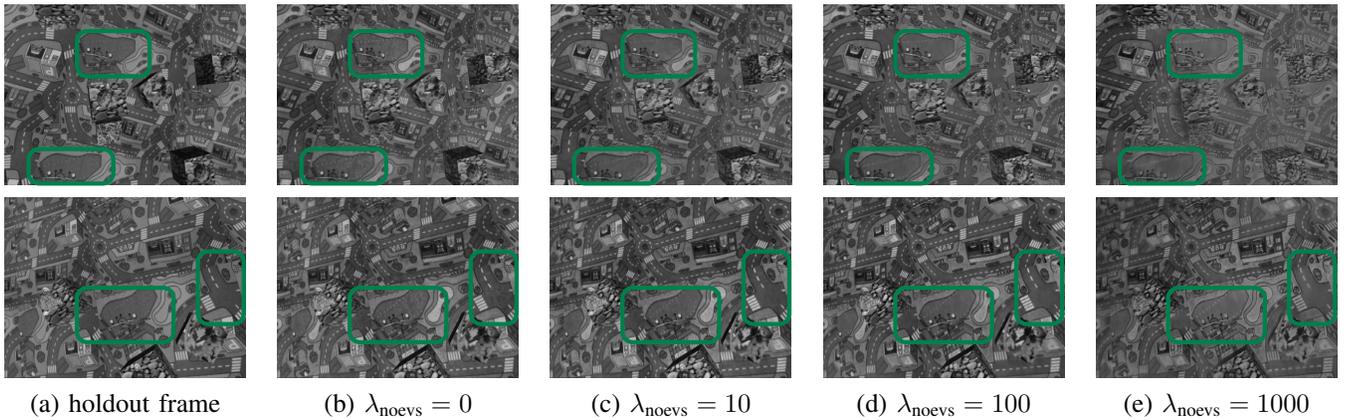

(a) holdout frame    (b) $\lambda_{\text{noevs}} = 0$    (c) $\lambda_{\text{noevs}} = 10$    (d) $\lambda_{\text{noevs}} = 100$    (e) $\lambda_{\text{noevs}} = 1000$

Fig. 1: The influence of the no-event loss $\mathcal{L}_{\text{noevs}}$ on the Shake Carpet 1 sequence. By increasing the weight $\lambda_{\text{noevs}}$ uniform brightness areas (which do not trigger events) are regularized and appear more smooth, such as in the landing ground or on the road network. We use $T_{\text{noevs}} = 20$ miliseconds to determine the no-event locations.

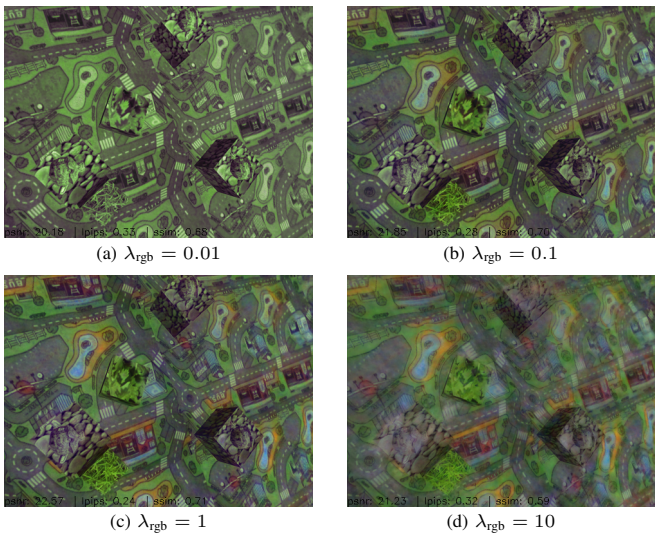

(a) $\lambda_{\text{rgb}} = 0.01$    (b) $\lambda_{\text{rgb}} = 0.1$

(c) $\lambda_{\text{rgb}} = 1$    (d) $\lambda_{\text{rgb}} = 10$

Fig. 2: Effect of $\lambda_{\text{rgb}}$ of the no-event loss on the scene Shake Carpet 1. This figure shows renderings from the same viewpoint as in Fig. 2 of the main paper (frame index 9).

| method | PSNR↑ | SSIM↑ | LPIPS↓ |
|---|---|---|---|
| E-NeRF | 16.15 | 0.60 | 0.145 |
| torch-ngp [3] | 27.03 | 0.86 | 0.077 |

TABLE II: Image quality metric on the scene mocap-1d-trans. Please note that E-NeRF (training with left event camera) is at a severe disadvantage here compared to NeRF (training with left grayscale camera), because both methods are evaluated in the left grayscale camera. There exists a baseline of around 4.49cm between grayscale and event cameras in TUMVIE [2], hence the pixel-wise image-quality metrics are heavily dependent on calibration errors.

| | torch-ngp (RGB) | E-NeRF (ours, event-only) |
|---|---|---|
| Training time | 0.21 | 0.29 |
| (i) Forward & backward pass | 99.28% | 97.40% |
| (ii) Load and sample events | 0.00% | 1.95% |
| (iii) Query Event Poses | 0.00% | 0.07% |
| (iv) Compute Rays | 0.61% | 0.50% |
| (v) Image Loading | 0.11% | 0.08% |

TABLE III: Training time in seconds per step. We use a batch size of 30096 rays on an NVIDIA A40 GPU. Runtime is measured with the time.perf_counter() function and averaged over the scenes Shake Carpet 1, Shake Moon 1 and Spiral 1. Logging is not included in the runtime measurement.

approximately the same training time. E-NeRF still has a slight overhead in the event sampling. We believe that this part of the algorithm can be considerably speed up by using more efficient data structures. Please note that runtime is not the focus of our paper.

*E. Details on the Encoding*

We use the hash grid encoding mechanism proposed by Mueller et al. [4]. This encoding has shown to lead to much faster convergence than the frequency-based encodings originally proposed in NeRF [1], while achieving similar quality [4], [5]. The employed encoding parameters in our paper are reported in Tab. IV.

In Tab. V, we perform an ablation over the encoding values on the scene Shake Carpet 1. If we change the number of levels $L$ from 16 to 14, we get a significant drop in PSNR from 26.65 to 24.99 (-1.66). On the other hand, if we set $L = 24$ we reach a PSNR value of 27.42 (+0.77). However, compared to our default setting, the computational time increases by a factor of around 2.24 when using $L = 24$. Hence we opt for a default value of $L = 16$ in all experiments.

As mentioned by Mueller et al. [4], another important hyperparameter is the finest resolution $N_{\max}$. In our implementation, we scale $N_{\max}$ per scene by $2048 \cdot B$, where $B$ is the bounding box size incorporating the scene ($B = 3$ on Shake Carpet 1). We change $N_{\max}$ to 512 and to 73728, respectively, and report the results in Tab. V. For $N_{\max} = 512$ we notice a peformance drop, whereas for $N_{\max} = 73728$ we only see a minor improvement.

*F. Depth results*

In general, the depth map quality of vanilla NeRF [1] is not perfect, especially if views are sparse [6], [7], [8], [9],

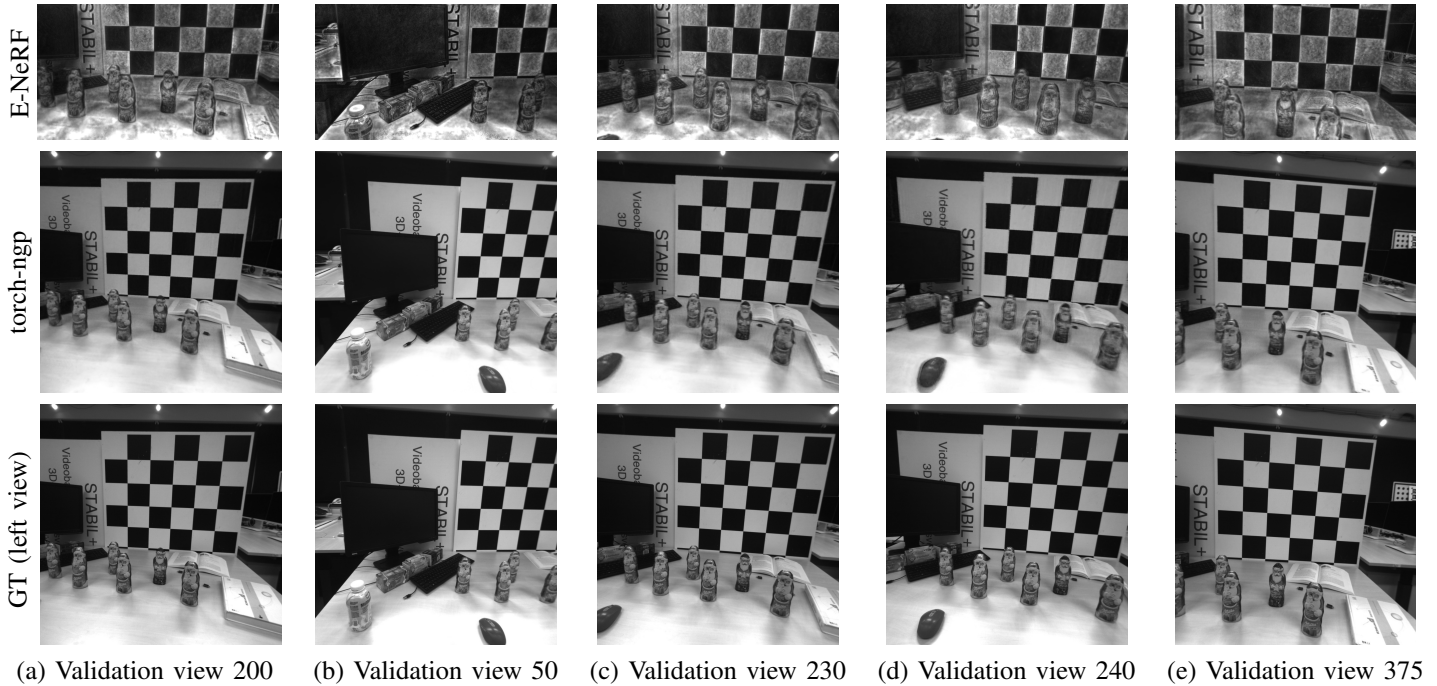

(a) Validation view 200  (b) Validation view 50  (c) Validation view 230  (d) Validation view 240  (e) Validation view 375

Fig. 3: Reconstruction results with slow camera motion and good illumination, using 90 training views for the torch-ngp baseline. See Tab. II for quantitative results.

| Parameter Name | Value |
| --- | --- |
| Number of levels $L$ | 16 |
| Dimensions per level $d$ | 2 |
| Log2 hashmap size $S$ | 19 |
| Coarsest Resolution $N_{min}$ | 16 |
| Finest Resolution $N_{max}$ | $B \cdot 2048$ |
| Bounding Box Size $B$ | 2 to 4[†] |

TABLE IV: Default hyperparameters for the employed hashgrid encoding [4]. [†]We set $B$ scene-depdent, see config files of the code release.

| Encoding Ablation | PSNR↑ | SSIM↑ | LPIPS↓ |
| --- | --- | --- | --- |
| default | 26.65 | 0.89 | 0.028 |
| $L = 14$ | 24.99 | 0.80 | 0.059 |
| $L = 24$ | 27.42 | 0.91 | 0.025 |
| $N_{max} = 512$ | 26.01 | 0.86 | 0.039 |
| $N_{max} = 72728$ | 26.81 | 0.89 | 0.030 |

TABLE V: Ablation study over the hashgrid encoding on the scene Shake Carpet 1. Default values for all other presented experiments are $L = 16$, $N_{max} = B \cdot 2048 = 6144$ ($B = 3$ for Shake Carpet 1).

| Method | Shake Moon 1 | Shake Moon 2 | Shake Carpet 1 | Spiral 1 | Spiral 2 | Avg. |
| --- | --- | --- | --- | --- | --- | --- |
| NeRF w/o blur | 1.28 | 1.19 | 0.78 | 1.67 | 1.99 | 1.38 |
| NeRF w/ blur | 1.40 | 1.28 | 0.90 | 1.68 | 2.05 | 1.46 |
| E2VID + NeRF | 1.40 | 1.33 | 0.84 | 1.76 | 2.07 | 1.48 |
| E-NeRF event-only | 1.29 | 1.22 | 0.84 | 1.62 | 1.99 | 1.39 |
| E-NeRF events & blur | 1.30 | 1.23 | 0.84 | 1.67 | 2.07 | 1.42 |

TABLE VI: Depth RMSE values in meters. We report the average over all available depth frames, i.e. over all train and test poses.

plain NeRF, see Tab. VI and Fig. 4. However, it should be straight-forward to extend our work by similar methods [6], [7], [8] in future work.

For further details, please refer to our released code under https://github.com/knelk/enerf.

[10], [11]. For example, the paper by Roessle et al. [6] shows the average depth RMSE is 1.163 meters on ScanNet [12], and 1.362 meters on Matterport3D [13], see Table 2 and Table 3 in [6]. Only if handled explicitly, the average RMSE can reach values in the 10-50 centimeter range. However, note that even if depth is handled explicitly, the qualitative maps show that texture and depth are still intertwined in the reconstructions, see Figure 3 in [6] in the 4th column (compared to ground truth). Since our work focuses on novel view quality, we do not include explicit depth regularizers. Hence, we get quantitative and qualitative results similar to

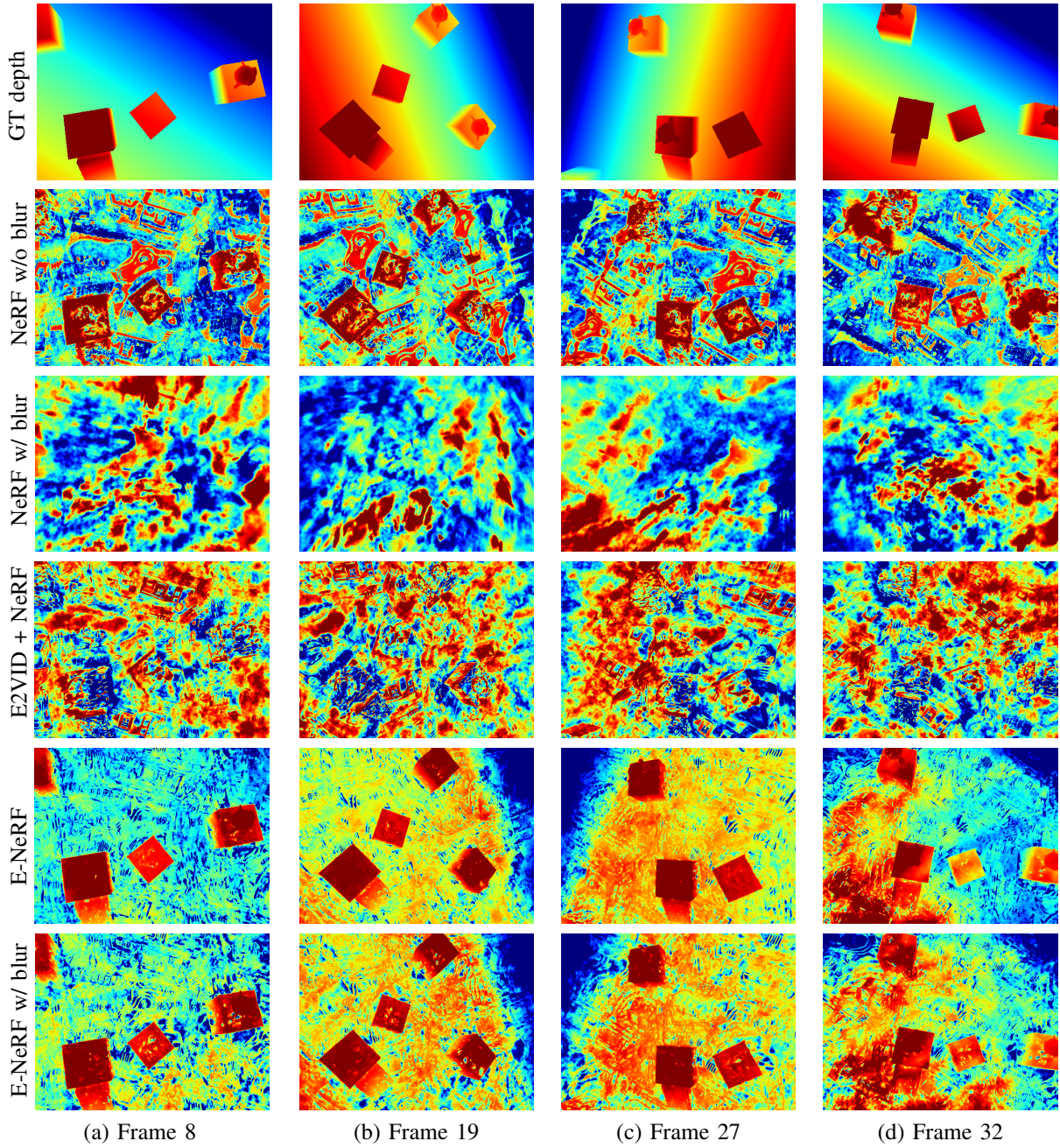

Fig. 4: Depth renderings from E-NeRF and various baseline methods on the sequence Shake Carpet 1 using 20 training views. The depth maps are still quite noisy. Due to spatially dense sampling of the event camera, the depth maps for E-NeRF (last two columns) show a less noisy and overall better depth result.



\end{document}

%% file: tables/spiral.tex
\bgroup
\setlength{\tabcolsep}{2pt}

\begin{table*}[tb]
\captionsetup{font=scriptsize}
\caption{SOTA comparison on the spiral sequences. The camera is moving on two different spiral paths (Spiral 1 and Spiral 2) around the synthetic scene, while continuously facing inward to the center of interest. The background texture shows a colored road network (see \autoref{fig::synt_motion_blur}).  The sparse sequences use fewer train views and all sequences show notable motion blur.
The proposed E-NeRF shows the best results in the event-only setting as well as when combining events and blurred frames. It shows even better results than NeRF trained on perfectly sharp images, which is partially due to the sparseness of the training views.
Deblur-NeRF is often worse than NeRF on the blurred frames because the blur is consistent (cf.\ Deblur-NeRF paper) and the views are sparse -- this reverses for some of the shake sequences in \autoref{tab::shake}.
The methods are evaluated on sharp test views using PSNR, SSIM~\cite{wang2004image}, and LPIPS~\cite{DBLP:conf/cvpr/ZhangIESW18}. The best and second-best result are marked in \textBF{bold} and \underline{underlined}, respectively. \textsuperscript{\textdagger}The "NeRF w/o blur" baseline uses groundtruth sharp frames that would not be available in real-world applications and is thus never marked. We use a batch size of 30096 event pairs in all spiral sequences.
}
\label{tab::spiral}

\begin{center}
\begin{tabular}{p{3.5cm} x{1cm} x{1cm} x{1cm} c @{\hspace{0.2cm}} x{1cm} x{1cm} x{1cm} c @{\hspace{0.2cm}} x{1cm} x{1cm} x{1cm} c @{\hspace{0.2cm}} x{1cm} x{1cm} x{1cm}}
\toprule
&
\multicolumn{3}{c}{Spiral 1} & &
\multicolumn{3}{c}{Spiral 2} & &
\multicolumn{3}{c}{Spiral 1 (sparse)} & &
\multicolumn{3}{c}{Spiral 1 (more-sparse)} \\[1.015ex]
&
\multicolumn{3}{c}{train/test views: 30/32} & &
\multicolumn{3}{c}{train/test views: 20/24} & &
\multicolumn{3}{c}{train/test views: 17/32} & &
\multicolumn{3}{c}{train/test views: \phantom{1}9/32} \\

\cmidrule{2-4}
\cmidrule{6-8}
\cmidrule{10-12}
\cmidrule{14-16}

&
\makecell{\notsotiny PSNR\,$\uparrow$}&
\makecell{\notsotiny SSIM\,$\uparrow$}&
\makecell{\notsotiny LPIPS\,$\downarrow$}& &
\makecell{\notsotiny PSNR\,$\uparrow$}&
\makecell{\notsotiny SSIM\,$\uparrow$}&
\makecell{\notsotiny LPIPS\,$\downarrow$}& &
\makecell{\notsotiny PSNR\,$\uparrow$}&
\makecell{\notsotiny SSIM\,$\uparrow$}&
\makecell{\notsotiny LPIPS\,$\downarrow$}& &
\makecell{\notsotiny PSNR\,$\uparrow$}&
\makecell{\notsotiny SSIM\,$\uparrow$}&
\makecell{\notsotiny LPIPS\,$\downarrow$}\\
\midrule
torch-ngp~\cite{torch-ngp} w/o blur\textsuperscript{\textdagger} & 25.74 & 0.74 & 0.06 & & 24.36 & 0.63 & 0.06 & & 20.67 & 0.38 & 0.11 & & 19.85 & 0.26 & 0.16 \\

torch-ngp~\cite{torch-ngp} w/\phantom{o} blur & 23.18 & 0.59 & 0.09 & & 23.85 & 0.54 & 0.09 & & 20.49 & 0.32 & 0.13 & & 19.80 & 0.26 & 0.17 \\

Deblur-NeRF~\cite{ma2021deblur} & 21.23 & 0.46 & 0.23 & & 18.10 & 0.18 & 0.72 & & 17.32 & 0.18 & 0.72 & & 16.88 & 0.16 & 0.70 \\

E2VID~\cite{DBLP:journals/pami/RebecqRKS21} + torch-ngp~\cite{torch-ngp} & 21.86 & 0.47 & 0.14 & & 21.83 & 0.30 & 0.13 & & 19.77 & 0.26 & 0.18 & & 19.75 & 0.26 & 0.17\\

E-NeRF event-only & \underline{26.91} & \underline{0.82} & \underline{0.04} & & \underline{25.71} & \underline{0.79} & \underline{0.04} & & \underline{26.91} & \underline{0.82} & \underline{0.04} & & \underline{26.91} & \underline{0.82} & \underline{0.04} \\

E-NeRF events \& blur & \textBF{28.23} & \textBF{0.85} & \textBF{0.04} & & \textBF{27.92} & \textBF{0.82} & \textBF{0.03} & & \textBF{27.34} & \textBF{0.85} & \textBF{0.04} & & \textBF{27.27} & \textBF{0.83} & \textBF{0.04} \\ 
\bottomrule
\end{tabular}
\vspace{-1.2\baselineskip}
\end{center}
\end{table*}

\egroup

%% file: tables/shake.tex
\bgroup
\setlength{\tabcolsep}{0pt}

\begin{table*}[tb]
\captionsetup{font=scriptsize}
\caption{SOTA comparison on the shake sequences. The camera is performing a random, forward-facing motion with abrupt changes of direction in each sequence. The background textures in the Shake Moon sequences show a gray moon landscape, whereas the background in Shake Carpet 1 consists of a colored road network (see \autoref{fig::synt_motion_blur}).
The proposed E-NeRF shows the best results in the event-only setting and outperforms NeRF trained on blurry frames. Deblur-NeRF is better than NeRF for mild blur, however, not for strong blur in sequence Shake Carpet 1 and not in \autoref{tab::spiral}. This shows that modelling the blur process can be beneficial, but using blur-resistant events is preferable and hence E-NeRF shows better results.
The methods are evaluated on sharp test views using PSNR, SSIM~\cite{wang2004image}, and LPIPS~\cite{DBLP:conf/cvpr/ZhangIESW18}. The best and second-best result are marked in \textBF{bold} and \underline{underlined}, respectively. \textsuperscript{\textdagger}The "NeRF w/o blur" baseline uses groundtruth sharp frames that would not be available in real-world applications and is thus never marked. We use a batch size of 20096 event pairs in all shake sequences}.
\label{tab::shake}

\begin{center}
\begin{tabular}{p{3.5cm} x{1.5cm} x{1.5cm} x{1.5cm} c @{\hspace{0.2cm}} x{1.5cm} x{1.5cm} x{1.5cm} c @{\hspace{0.2cm}} x{1.5cm} x{1.5cm} x{1.5cm}}
\toprule
& 
\multicolumn{3}{c}{Shake Moon 1} & & 
\multicolumn{3}{c}{Shake Moon 2} & & 
\multicolumn{3}{c}{Shake Carpet 1} \\[1.015ex]

&
\multicolumn{3}{c}{train/test views: 39/38, mild blur} & &
\multicolumn{3}{c}{train/test views: 33/32, medium blur} & &
\multicolumn{3}{c}{train/test views: 20/21, strong blur} \\

\cmidrule{2-4}
\cmidrule{6-8}
\cmidrule{10-12}

&
\makecell{\notsotiny PSNR\,$\uparrow$}&
\makecell{\notsotiny SSIM\,$\uparrow$}&
\makecell{\notsotiny LPIPS\,$\downarrow$}& &
\makecell{\notsotiny PSNR\,$\uparrow$}&
\makecell{\notsotiny SSIM\,$\uparrow$}&
\makecell{\notsotiny LPIPS\,$\downarrow$}& &
\makecell{\notsotiny PSNR\,$\uparrow$}&
\makecell{\notsotiny SSIM\,$\uparrow$}&
\makecell{\notsotiny LPIPS\,$\downarrow$}\\
\midrule

torch-ngp~\cite{torch-ngp} w/o blur\textsuperscript{\textdagger} & 33.12 & 0.93 & 0.02 & & 28.41 & 0.84 & 0.03 & & 29.42 & 0.87 & 0.03 \\

torch-ngp~\cite{torch-ngp} w/\phantom{o} blur & 24.22 & 0.61 & 0.12 & & 22.04 & 0.52 & 0.13 & & 19.69 & 0.31 & 0.22 \\

Deblur-NeRF~\cite{ma2021deblur} & 25.97 & 0.68 & 0.20 && 24.78 & 0.58 & 0.36 && 16.44 & 0.18 & 0.69 \\

E2VID~\cite{DBLP:journals/pami/RebecqRKS21} + torch-ngp~\cite{torch-ngp} & 22.39 & 0.80 & 0.05 && 22.31 & 0.78 & 0.06 && 22.81 & 0.80 & 0.05 \\

E-NeRF event-only & \underline{31.10} & \underline{0.94} & \underline{0.02} && \underline{30.32} & \underline{0.93} & \underline{0.02} && \underline{26.65} & \textBF{0.89} & \underline{0.03} \\

E-NeRF events \& blur & \textBF{31.35} & \textBF{0.95} & \textBF{0.01} && \textBF{30.60} & \textBF{0.93} & \textBF{0.02} && \textBF{27.90} & \underline{0.89} & \textBF{0.03} \\
\bottomrule
\end{tabular}
\vspace{-1.0\baselineskip}
\end{center}
\end{table*}

\egroup

%% file: tables/ablation.tex
\bgroup
\setlength{\tabcolsep}{0pt}

\begin{table}[tb]
\captionsetup{font=scriptsize}
\caption{Ablation study of E-NeRF. We (i)~replace the loss function \autoref{eq:evModel2} by \autoref{eq:evModel3}, (ii)~change the event sampling from an average event pair distance of 1ms to 30ms, (iii)~add the no-event loss from \autoref{eq:noevModel} to the loss function by sampling up to one fourth of no-events in each batch (with $T_{\text{noevs}} = 25$ms). While those modifications do not improve the average metrics, we find that using them on real data can result in better visual quality.
The methods are evaluated using PSNR, SSIM~\cite{wang2004image}, and LPIPS~\cite{DBLP:conf/cvpr/ZhangIESW18}. The best and second-best result are marked in \textBF{bold} and \underline{underlined}, respectively.}
\label{tab::ablation}

\begin{center}
\begin{tabular}{p{0.8cm} p{4.0cm} x{1cm} x{1cm} x{1cm}}
\toprule
& &
\makecell{\notsotiny PSNR\,$\uparrow$}&
\makecell{\notsotiny SSIM\,$\uparrow$}&
\makecell{\notsotiny LPIPS\,$\downarrow$}\\
\midrule
& E-NeRF  & \textBF{28.14} & \textBF{0.88} & \textBF{0.028}  \\   
 (i) & normalized loss (\autoref{eq:evModel3})  & 28.07 & 0.86 & 0.031  \\ 
 (ii) & event sampling 60ms window & 28.13 & 0.86 & 0.033 \\
 (iii) & no-event loss $\lambda_\text{noevs}=1$  & 27.76 & 0.87 & \underline{0.030} \\
\bottomrule

\end{tabular}
\end{center}
\vspace{-0.3cm}
\end{table}

\egroup

%% file: tables/noEvsReal.tex
\begin{figure*}
	\centering
    \begin{tabular}{c c c c c}
    \begin{tikzpicture} 
        \node[anchor=south west,inner sep=0] (image) at (0,0) {\includegraphics[width=0.18\textwidth]{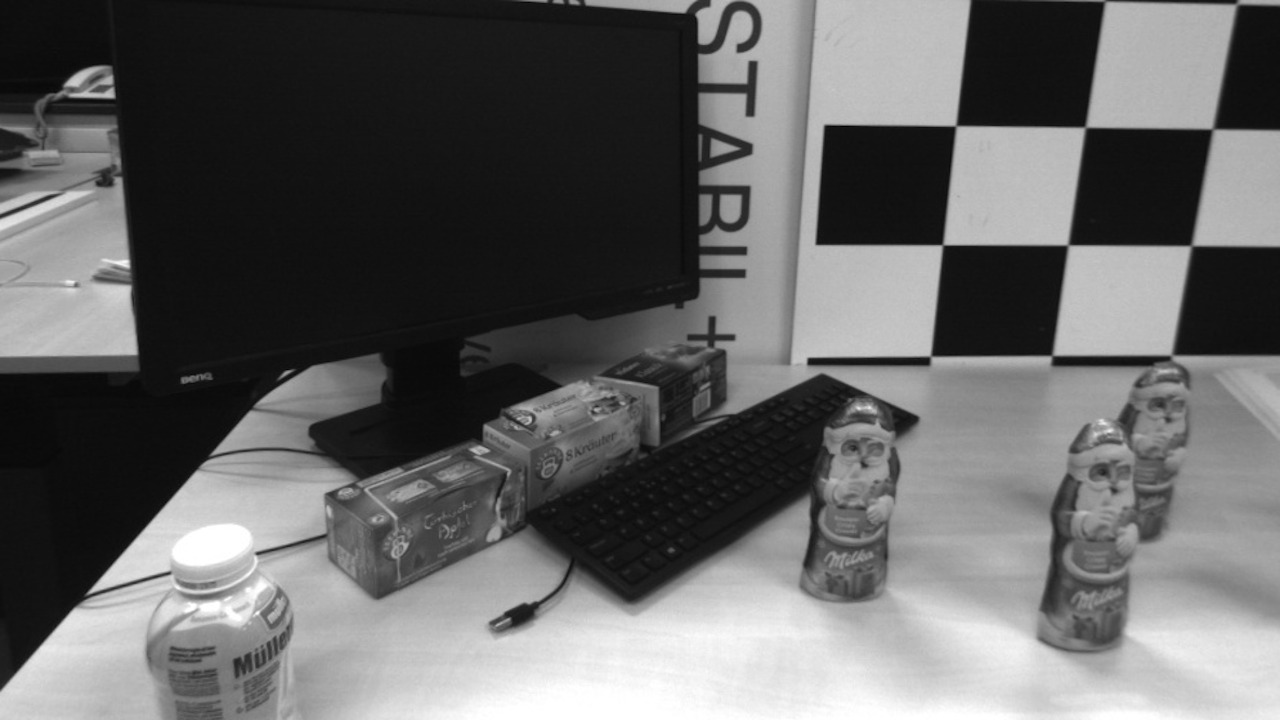}};
        \begin{scope}[x={(image.south east)},y={(image.north west)}]
            \draw[mycolor,ultra thick,rounded corners] (0.3,0.01) rectangle (0.7,0.2);
            \draw[mycolor,ultra thick,rounded corners] (0.6,0.5) rectangle (0.99,0.99);
        \end{scope}
    \end{tikzpicture}    
    &     
    \begin{tikzpicture} 
        \node[anchor=south west,inner sep=0] (image) at (0,0) {\includegraphics[width=0.18\textwidth]{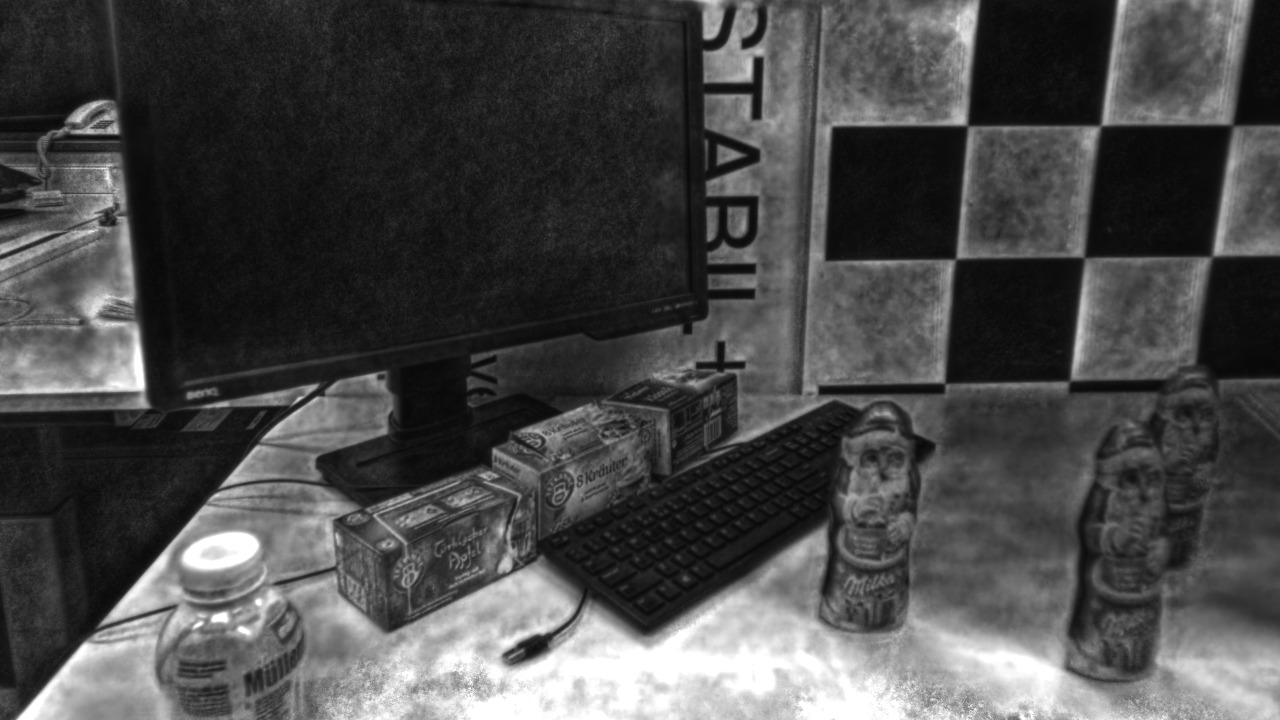}};
        \begin{scope}[x={(image.south east)},y={(image.north west)}]
            \draw[mycolor,ultra thick,rounded corners] (0.3,0.01) rectangle (0.7,0.2);
            \draw[mycolor,ultra thick,rounded corners] (0.6,0.5) rectangle (0.99,0.99);
        \end{scope}
    \end{tikzpicture} 
    & 
    \begin{tikzpicture} 
        \node[anchor=south west,inner sep=0] (image) at (0,0) {\includegraphics[width=0.18\textwidth]{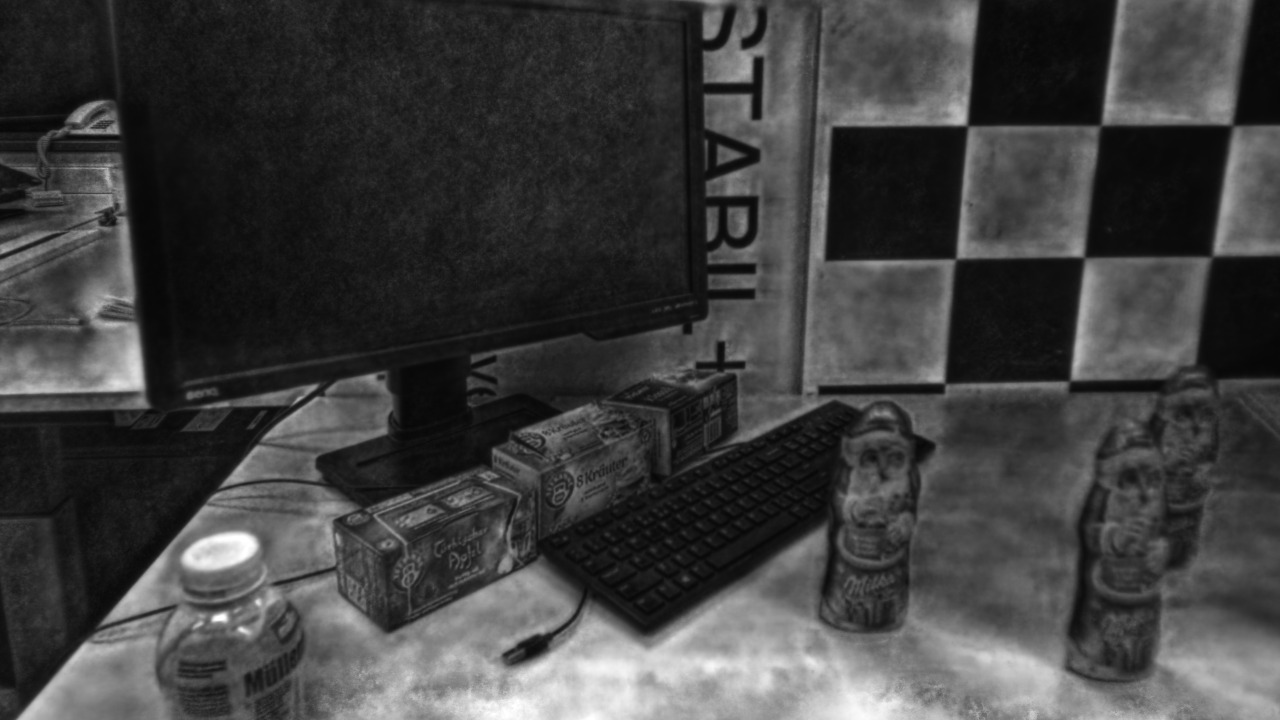}};
        \begin{scope}[x={(image.south east)},y={(image.north west)}]
            \draw[mycolor,ultra thick,rounded corners] (0.3,0.01) rectangle (0.7,0.2);
            \draw[mycolor,ultra thick,rounded corners] (0.6,0.5) rectangle (0.99,0.99);
        \end{scope}
    \end{tikzpicture} 
    &
    \begin{tikzpicture} 
        \node[anchor=south west,inner sep=0] (image) at (0,0) {\includegraphics[width=0.18\textwidth]{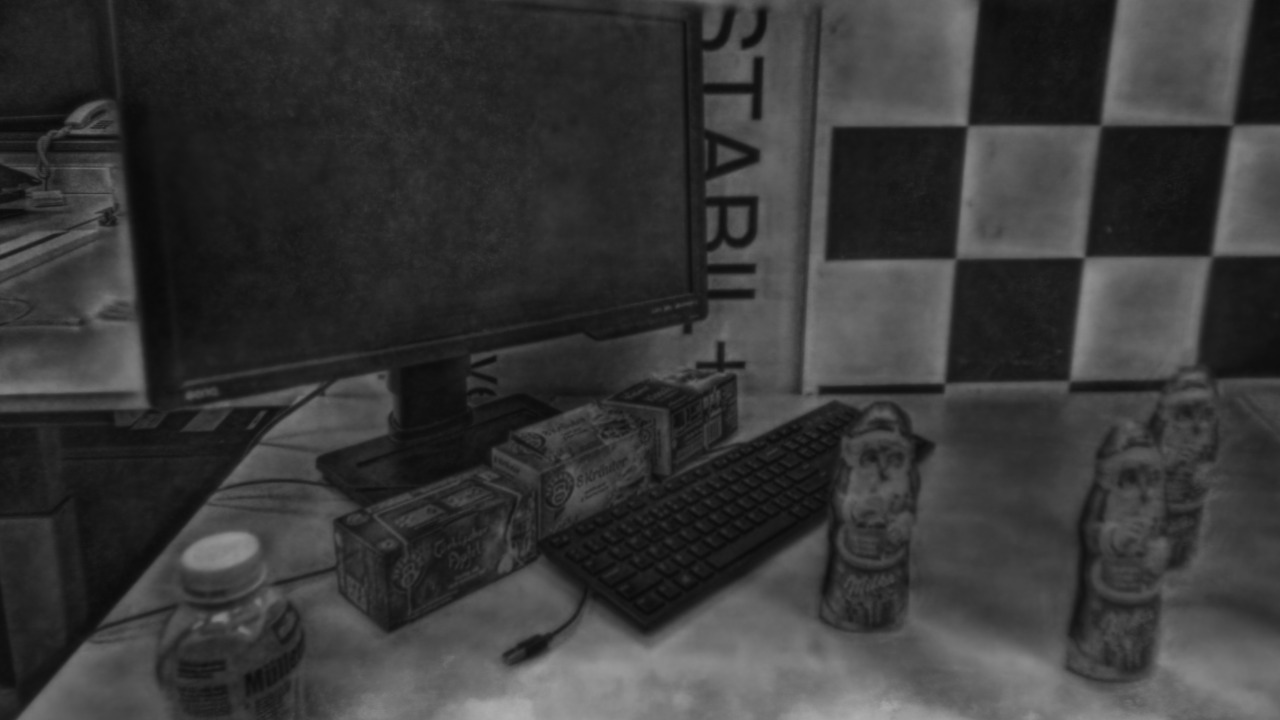}};
        \begin{scope}[x={(image.south east)},y={(image.north west)}]
            \draw[mycolor,ultra thick,rounded corners] (0.3,0.01) rectangle (0.7,0.2);
            \draw[mycolor,ultra thick,rounded corners] (0.6,0.5) rectangle (0.99,0.99);
        \end{scope}      
    \end{tikzpicture} 
    & 
    \begin{tikzpicture} 
        \node[anchor=south west,inner sep=0] (image) at (0,0) {\includegraphics[width=0.18\textwidth]{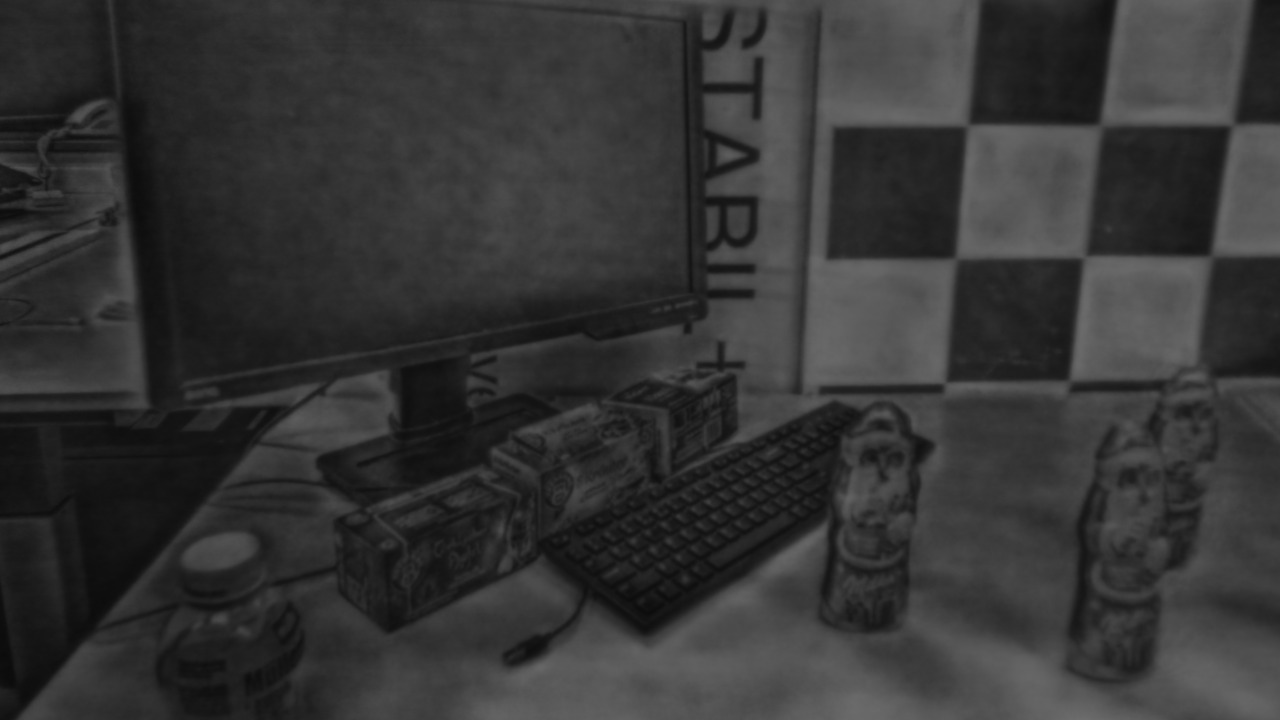}};
        \begin{scope}[x={(image.south east)},y={(image.north west)}]
            \draw[mycolor,ultra thick,rounded corners] (0.3,0.01) rectangle (0.7,0.2);
            \draw[mycolor,ultra thick,rounded corners] (0.6,0.5) rectangle (0.99,0.99);
        \end{scope}      
    \end{tikzpicture} 
    \\[0.2ex]
    \begin{tikzpicture} 
        \node[anchor=south west,inner sep=0] (image) at (0,0) {\includegraphics[width=0.18\textwidth]{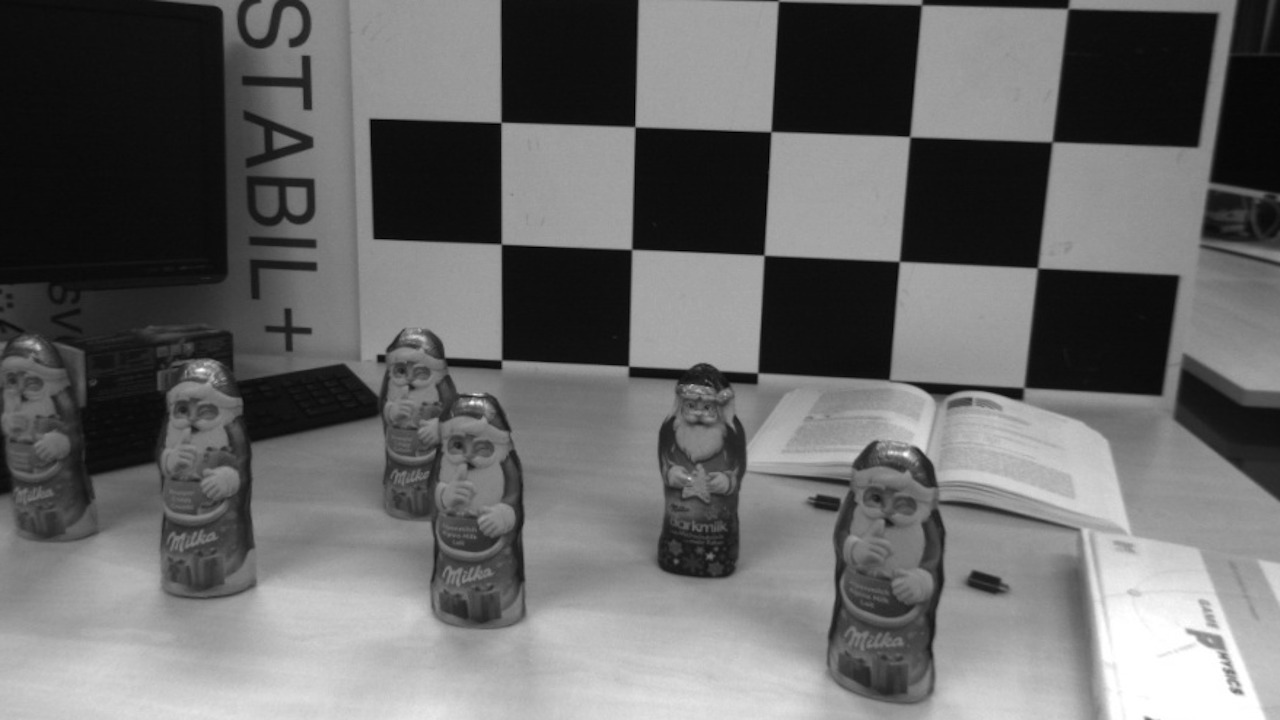}};
        \begin{scope}[x={(image.south east)},y={(image.north west)}]
            \draw[mycolor,ultra thick,rounded corners] (0.27,0.45) rectangle (0.95,0.99);
        \end{scope}
    \end{tikzpicture}    
    &     
    \begin{tikzpicture} 
        \node[anchor=south west,inner sep=0] (image) at (0,0) {\includegraphics[width=0.18\textwidth]{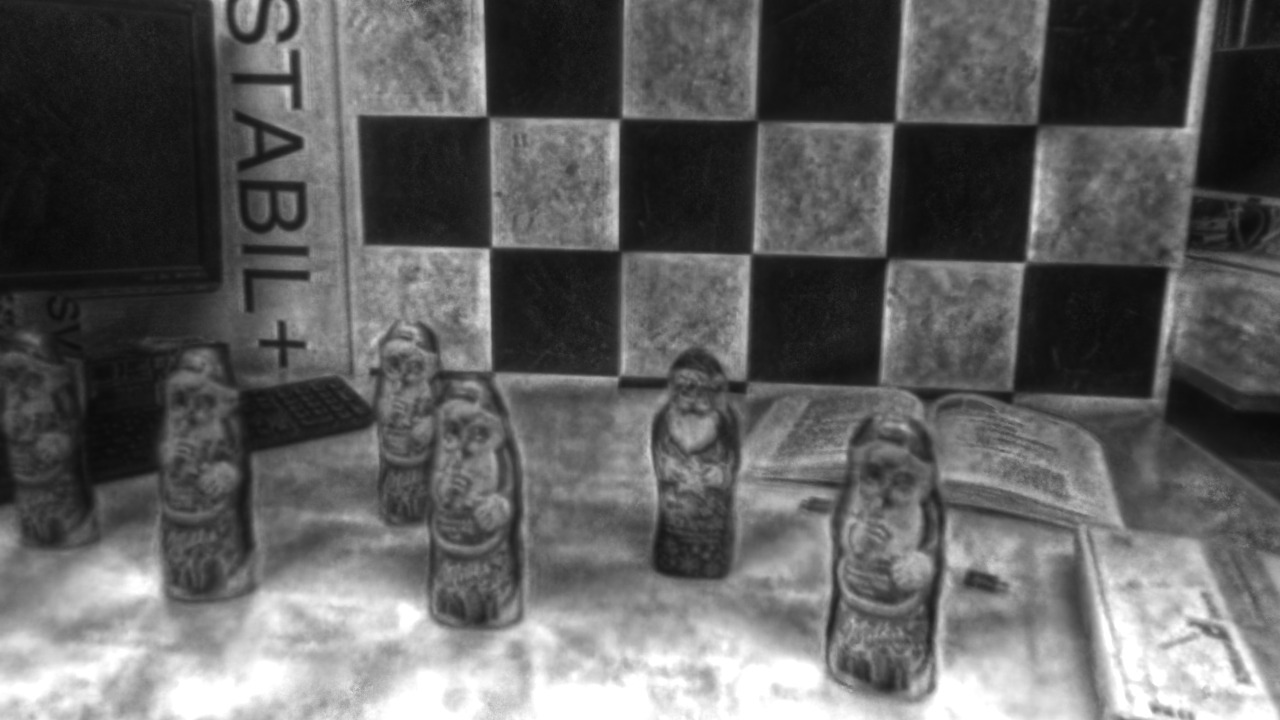}};
        \begin{scope}[x={(image.south east)},y={(image.north west)}]
            \draw[mycolor,ultra thick,rounded corners] (0.27,0.45) rectangle (0.95,0.99);
        \end{scope}
    \end{tikzpicture} 
    & 
    \begin{tikzpicture} 
        \node[anchor=south west,inner sep=0] (image) at (0,0) {\includegraphics[width=0.18\textwidth]{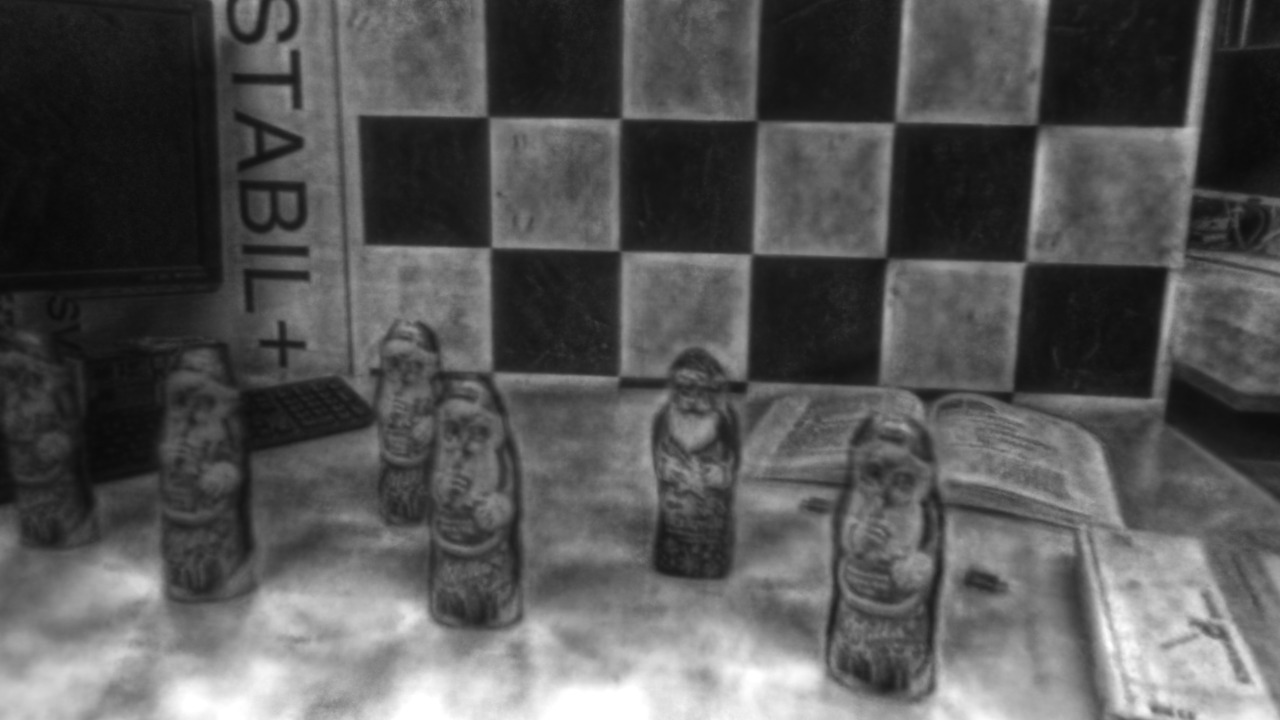}};
        \begin{scope}[x={(image.south east)},y={(image.north west)}]
            \draw[mycolor,ultra thick,rounded corners] (0.27,0.45) rectangle (0.95,0.99);
        \end{scope}
    \end{tikzpicture} 
    &
    \begin{tikzpicture} 
        \node[anchor=south west,inner sep=0] (image) at (0,0) {\includegraphics[width=0.18\textwidth]{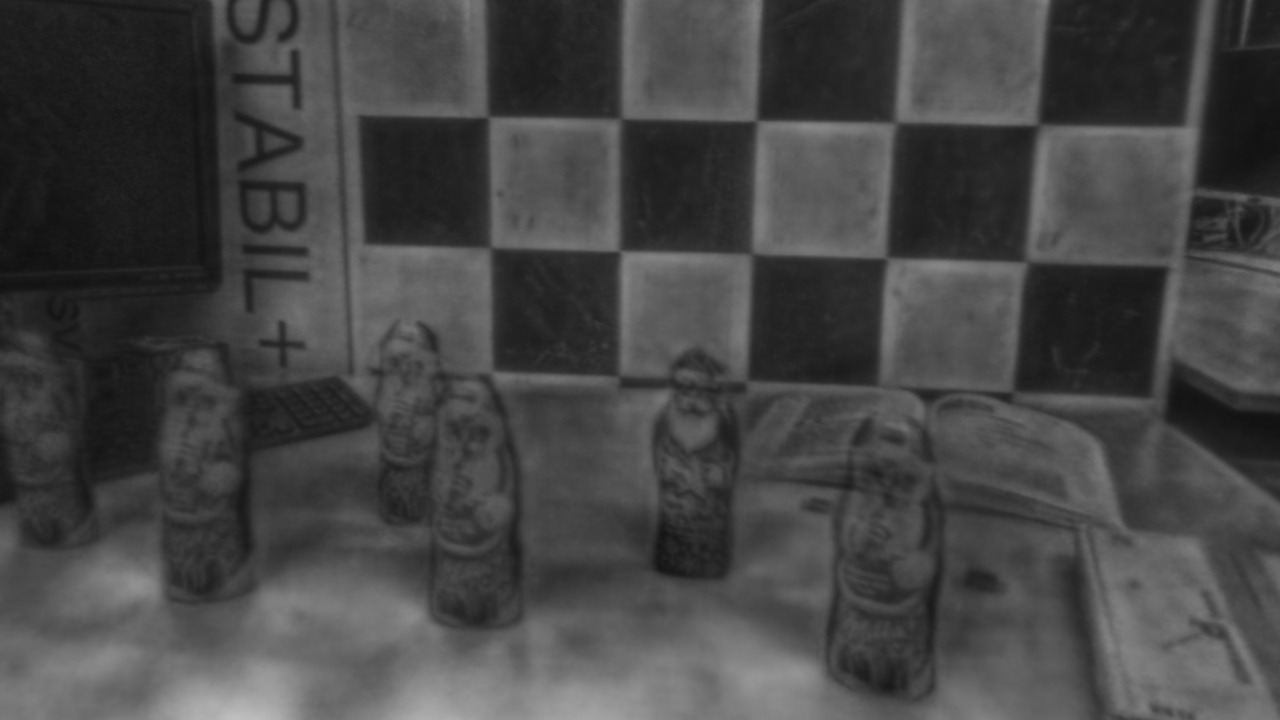}};
        \begin{scope}[x={(image.south east)},y={(image.north west)}]
            \draw[mycolor,ultra thick,rounded corners] (0.27,0.45) rectangle (0.95,0.99);
        \end{scope}      
    \end{tikzpicture} 
    & 
    \begin{tikzpicture} 
        \node[anchor=south west,inner sep=0] (image) at (0,0) {\includegraphics[width=0.18\textwidth]{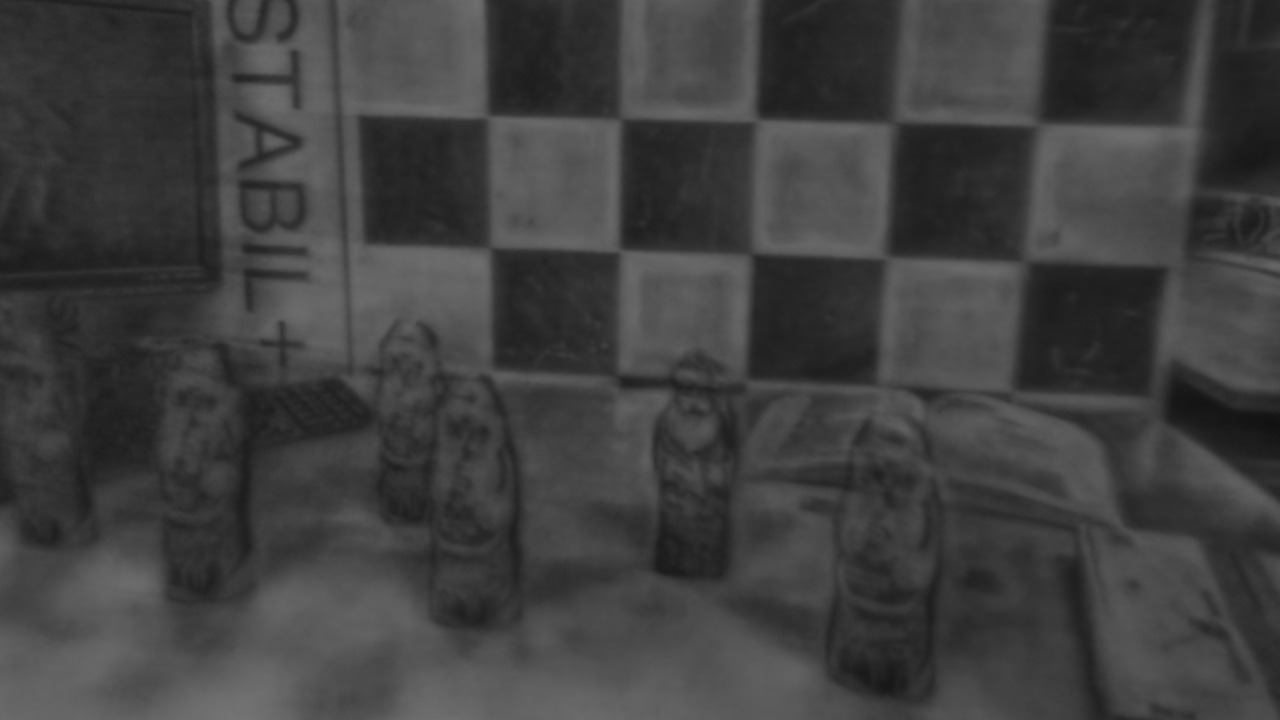}};
        \begin{scope}[x={(image.south east)},y={(image.north west)}]
            \draw[mycolor,ultra thick,rounded corners] (0.27,0.45) rectangle (0.95,0.99);
        \end{scope}      
    \end{tikzpicture} 
    \\[0.2ex]
    (a) holdout frame & (b) $\lambda_\text{noevs}=0$ & (c) $\lambda_\text{noevs}=10$ & (d) $\lambda_\text{noevs}=100$ & (e) $\lambda_\text{noevs}=1000$\\
    \end{tabular}
\caption{The influence of the no-event loss $\mathcal{L}_\text{noevs}$ on the sequence mocap-1d-trans \cite{DBLP:conf/iros/KlenkCDC21}, where the camera is moving slowly and the illumination is good. By increasing the weight $\lambda_\text{noevs}$, uniform brightness areas which do not trigger events are regularized and appear more smooth, such as the checkerboard in the background or the table surface. We use $T_\text{noevs}=20$ miliseconds to determine the no-event locations and a batch size of 20096 event pairs.}
\label{fig::noEvsReal}
\end{figure*}
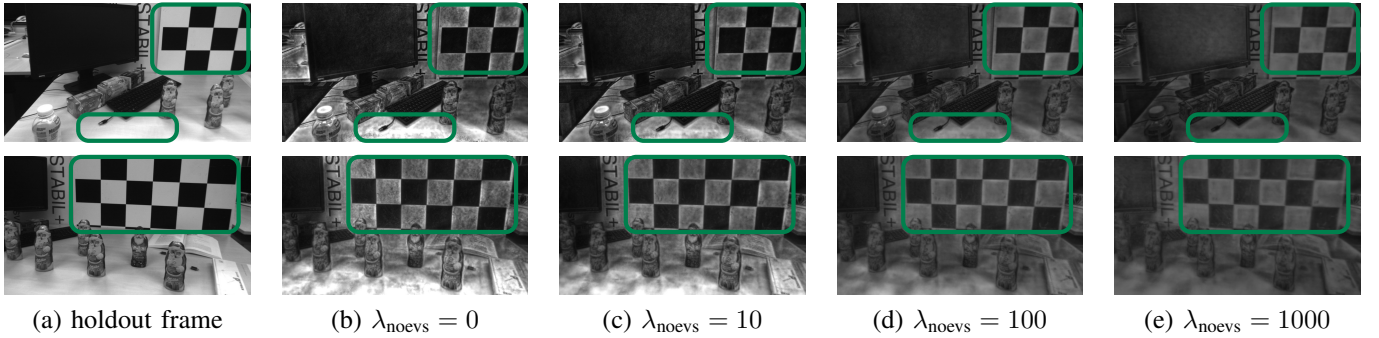\textbf{}